# Semi-supervised classification of radiology images with NoTeacher: A Teacher that is not Mean

Balagopal Unnikrishnan[a,1,*], Cuong Nguyen[a,1], Shafa Balaram[a,b], Chao Li[a], Chuan Sheng Foo[a], Pavitra Krishnaswamy[a,*]

[a]*Institute for Infocomm Research, Agency for Science Technology and Research (A*STAR), Singapore*
[b]*National University of Singapore, Singapore*

**Abstract**

Deep learning models achieve strong performance for radiology image classification, but their practical application is bottlenecked by the need for large labeled training datasets. Semi-supervised learning (SSL) approaches leverage small labeled datasets alongside larger unlabeled datasets and offer potential for reducing labeling cost. In this work, we introduce NoTeacher, a novel consistency-based SSL framework which incorporates probabilistic graphical models. Unlike Mean Teacher which maintains a teacher network updated via a temporal ensemble, NoTeacher employs two independent networks, thereby eliminating the need for a teacher network. We demonstrate how NoTeacher can be customized to handle a range of challenges in radiology image classification. Specifically, we describe adaptations for scenarios with 2D and 3D inputs, uni and multi-label classification, and class distribution mismatch between labeled and unlabeled portions of the training data. In realistic empirical evaluations on three public benchmark datasets spanning the workhorse modalities of radiology (X-Ray, CT, MRI), we show that NoTeacher achieves over 90-95% of the fully supervised AUROC with less than 5-15% labeling budget. Further, NoTeacher outperforms established SSL methods with minimal hyperparameter tuning, and has implications as a principled and practical option for semi-supervised learning in radiology applications.

## 1. Introduction

Deep learning approaches offer state-of-the-art performance for several image classification applications in radiology. Recent successes include chest radiograph diagnosis, brain tumor prognostication, fracture detection, and breast cancer screening (Rajpurkar et al., 2018; Lao et al., 2017; Lindsey et al., 2018; McKinney et al., 2020). However, these efforts typically require large labeled training datasets assembled through resource-intensive labeling by specialized domain experts. Further, as discordance between expert raters can result in noisy labels, even more resource-intensive consensus rating across multiple blinded raters is often required (Langlotz et al., 2019). Moreover, as models developed with a dataset from one center or cohort may often not generalize well to datasets from other centers or cohorts, continuing annotation of diverse datasets from different centers or populations is essential to ensure practical relevance Choudhary et al. (2020). Hence, annotation burden is a major bottleneck to developing and translating models for image classification applications in radiology.

Semi-supervised learning (SSL) approaches have potential to significantly lower the data annotation barrier while still maintaining the performance levels of a fully

*Correspondence: Balagopal_Unnikrishnan@i2r.a-star.edu.sg; Nguyen_Manh_Cuong@i2r.a-star.edu.sg; pavitrak@i2r.a-star.edu.sg.
+65-6408-2450 (Tel.); +65-6776-1378 (Fax).
[1]Equal Contribution.

supervised learning paradigm (Van Engelen and Hoos, 2020). This is because SSL approaches leverage small numbers of labeled images alongside larger numbers of unlabeled images for model development. Given that clinical databases often contain large amounts of unlabeled data, such approaches present important opportunities for leveraging existing information and minimizing cost for model development.

SSL approaches have been widely employed in the computer vision community for applications involving classification of natural scene images (Rebuffi et al., 2020). However, the problem of classifying images in radiology (X-Ray, CT, MR) is quite distinct from that of classifying photographic natural scene images. First, images in radiology typically span beyond 2D to 3D given the increasing prevalence of cross-sectional imaging; and are often classified based on fine-grained or subtle features. Second, the classification task can often involve detection of multiple abnormality types in the same image. Third, when the learning paradigm employs a mix of labeled and unlabeled data, the radiology setting imposes practical constraints that preclude ideal alignment of class distributions across the labeled and unlabeled subsets. These differences have rendered it challenging to demonstrate a unified SSL approach that can cater to the diverse challenges for the range of radiology image classification needs.

The best performing SSL approaches on computer vision benchmarks are typically consistency-based; these encourage a classifier's predictions to be consistent with a target on the unlabeled data. For instance, the popular Mean Teacher (MT) method (Tarvainen and Valpola, 2017) enforces consistency between predictions from two networks, termed student and teacher, where the teacher is a time-averaged version of student and is used for inference at test time. The medical imaging community has attempted to leverage these methods (especially Mean Teacher) for medical image analysis (Su et al., 2019). Further, recent efforts have proposed variants of Mean Teacher that exploit unlabeled data by modeling semantic relations between samples (Liu et al., 2020) and employ uncertainty maps to improve the teacher model's prediction (Yu et al., 2019). However, Mean Teacher and its variants fundamentally rely on a time-averaged student as a consistency target (teacher). This essentially leads the model to enforce self-consistency, and renders the approach vulnerable to confirmation bias (Ke et al., 2019; Pham et al., 2020).

To address the above challenges, we propose a means to remove the reliance on the time-averaging component by introducing the NoTeacher semi-supervised learning framework. Our framework, which employs graphical models to enforce prediction consistency, not only overcomes the drawbacks of Mean Teacher but is also customizable to handle the unique challenges of radiology applications. Our main contributions are summarized as follows:

1. We introduce a novel semi-supervised learning framework, NoTeacher, for radiology image classification. NoTeacher employs graphical models to enhance consistency and reduce confirmation bias in relation to Mean Teacher.
2. We describe ways to customize the NoTeacher framework to cater to distinct types of classification tasks spanning 2D and/or 3D inputs associated with single and/or multiple labels.
3. We present extensions of the NoTeacher framework for SSL scenarios where there is a mismatch between the class distributions of the labeled and unlabeled training datasets.
4. We connect NoTeacher to theoretically principled multi-view and co-training approaches, and empirically demonstrate these connections via analysis of model evolution during semi-supervised training.
5. We design realistic semi-supervised experiments based on considerations of practical annotation workflows, and evaluate our NoTeacher framework on 2D and 3D datasets (Rajpurkar et al., 2017; Flanders et al., 2020; Bien et al., 2018) spanning the three workhorse modalities of radiology (X-Ray, CT and MRI). We demonstrate that NoTeacher outperforms established baseline methods with minimal hyperparameter tuning, and achieves over 90-95% of the fully supervised AUROC with less than 5-15% labeling budget (depending on the dataset).

This paper constitutes an extended version of Unnikrishnan et al. (2020). We focus on expansions of the NoTeacher (NoT) framework to enhance practical relevance for a broader range of radiology image classification tasks. We introduce extensions of NoT for 3D scans and for cases with distinct class distributions in the labeled and unlabeled datasets. We then perform rigorous and extended evaluations on two radiograph datasets to specifically demonstrate gains for (a) multi-label cases and (b) labeled-unlabeled class distribution mismatch scenarios. We also extend beyond radiography to magnetic resonance imaging (MRI) by including evaluations for a new 3D MR dataset. Finally, we anchor the NoTeacher framework within the theoretically grounded context of co-training approaches. We also present empirical analyses of NoT performance on the three datasets to demonstrate the effectiveness of the proposed mechanism for enforcing prediction consistency. Our results show that NoTeacher enables strong performance gains over the prevailing state-of-the-art SSL methods across the three common radiology modalities (X-Ray, CT and MR). As NoTeacher is a theoretically principled framework amenable to customization for a range of practically relevant scenarios, it has implications for a range of radiology image analysis tasks.

## 2. Related Work

### 2.1. Semi-supervised Learning Methods

Relevant semi-supervised learning (SSL) methods can be broadly classified into pseudo-labeling, consistency-based, adversarial learning and graph-based methods. Below, we review key works from each category and associated efforts in the medical imaging literature. We provide a review of methods that are used recently. For a more comprehensive review, we refer to (Van Engelen and Hoos, 2020).

**Pseudo-labeling methods:** Pseudo-labeling (PSU) (Lee, 2013) is amongst the most simple and commonly used semi-supervised algorithms. It involves two steps: (a) training a supervised model with a labeled subset of the training data, and (b) using this supervised model to infer pseudo-labels for the unlabeled portion of the training data. The second step employs the combined data and recalculates the pseudo-labels at every weight update to inform further the loss function of the first supervised learning task. It is effectively equivalent to entropy regularization (Rizve et al., 2021) and works well for cases with low-density separation between classes. However, it can also reinforce errors learnt by the initial model (Arazo et al., 2020).

**Adversarial learning methods:** Virtual Adversarial Training (VAT) (Miyato et al., 2018) represents another line of research that seeks to utilize adversarial examples for semi-supervised learning. It employs a regularization that trains the distribution of the target label (conditional on input features) to be isotropically smooth around each input data point. Specifically, for each sample, it identifies a virtual adversarial perturbation that most greatly alters the current output distribution. This step is unsupervised and only requires the features associated with the sample. The robustness of the model against the adversarial perturbation is defined by a local distributional smoothness (LDS) loss. During training, VAT optimizes model parameters to minimize (a) the negative log-likelihood for the labeled dataset and (b) the average LDS between the unperturbed and perturbed output distributions over all input data. Although VAT does not explicitly generate adversarial samples, it identifies the adversarial perturbations that would affect the model performance and smooths the model in those vulnerable directions. This category of SSL methods relies heavily on the generation of extra training samples, and takes longer training cycles due to the extra forward/backward passes required to compute the perturbation. A few methods have applied adversarial learning to medical imaging, including a domain adaptation method for endoscopy (Mahmood et al., 2018) and a classification method for dermatology (Margeloiu et al., 2020).

**Graph-based methods:** These methods build a graph in which each node represents a sample, while the graph edges connects nodes that are likely to have the same label. Then, a learning function such as label propagation (Zhu and Ghahramani, 2002; Zhou et al., 2004) or a constraint such as manifold regularization (Belkin et al., 2006) is applied over the graph to deduce the label for unlabeled samples. While graph-based methods have the intrinsic power to leverage the underlying structure of the data, they often rely on strong assumptions (e.g., label-smoothness assumption or manifold assumption), and are transductive in nature – hence limiting general scalability. In the area of medical imaging, GraphX[NET] (Aviles-Rivero et al., 2019) is a label-propagating graph-based method which was developed for X-Ray classification. However, it does not support the multi-label setting, was not demonstrated on 3D scans and is yet to be compared to other SSL methods.

**Consistency-based methods:** This line of work is most related to our study. Consistency-based methods are built upon the assumption that model prediction should be consistent between different augmentations. Therefore, consistency regularization can be applied to unlabeled data as an additional constraint. Mean Teacher (MT) (Tarvainen and Valpola, 2017) is the dominant consistency-based method. MT employs two networks with similar architecture: a student model and a teacher model. Given a batch of semi-supervised training data, MT computes a total loss which is defined as a weighted sum of a supervised term and a consistency term. The supervised term is calculated between the student model's prediction and the target label, while the consistency term is between the posterior outputs of the two networks. The student model backpropagates directly using gradients from the total loss. In the meantime, the teacher model is updated via computing an exponential moving average (EMA) over the parameters of the student network. Essentially, the teacher model is a temporal ensemble of the student model in the parameter space.

In the medical imaging domain, several studies have applied and adapted Mean Teacher for segmentation tasks. Recent efforts have proposed two notable variants of MT. Liu et al. (2020) proposed Sample-Relation-Consistency Mean Teacher (SRC-MT). Apart from the vanilla consistency regularization, this approach enforces the consistency of semantic relations on training samples under random perturbations, in order to exploit additional semantic information from unlabeled data. This method was demonstrated on 2D classification tasks. Yu et al. (2019) developed Uncertainty-Aware Mean Teacher (UA-MT) which employs an uncertainty map to improve the teacher model's prediction.

MT holds an edge over supervised learning when the teacher generates better expected targets, or pseudo-labels, to train its student. However, because the teacher model is basically a temporal ensemble of the student model in the parameter space, MT has several potential drawbacks. First, it has been shown that promoting the consistency between the student model and its historical self may lead to confirmation bias or unwanted propagation of label noise (Ke et al., 2019; Pham et al., 2020). Second, the teacher model is known to be sensitive to the choice of the EMA hyperparameter, especially in realistic evaluation regimes (Oliver et al., 2018). If the EMA hyperparameter is set outside of an ideal narrow range, performance of MT can degrade quickly. Finally, since MT does not specify a systematic method to determine the consistency weight, it is unclear as to how we may tune hyperparameters for varied datasets. Therefore, recent variants of MT have resorted to other consistency regularization schemes (Perone and Cohen-Adad, 2018; Su et al., 2019), but they still suffer from the inherent drawbacks of MT.

**Other SSL Methods:** Aside from the above categories, there are a few other notable SSL methods. For example, generative methods learn the underlying distribution from unlabeled data via generative adversarial networks (GANs) (Salimans et al., 2016). But these do not easily scale to high-resolution images typical in radiology (Madani et al., 2018b,a) and patch-wise adaptations (Lecouat et al., 2018) neglect the necessary global context. Further, MixMatch (Berthelot et al., 2019) and Gyawali et al. (2020) emphasizes data augmentation. MixMatch creates extra training samples via interpolating between labeled and unlabeled samples in an augmentation step called MixUp. This interpolation is performed not only in the input space but also in the target label space, to generate a diverse training set. However, the MixUp augmentation technique cannot be easily applied to various types of medical images such as 3D MR scans. Gyawali et al. (2019) combined temporal ensembling and disentangled representation learning to improve multi-label classification of chest X-Ray. This method, however, requires a two-stage training procedure and cannot be trained end-to-end.

*2.2. Specific Challenges in Medical Imaging*

We further consider related works that address the specific challenges of class imbalance and 3D image handling, that are relevant to our study.

**SSL for class imbalanced medical imaging applications:** To handle class imbalance, some recent work such as Wang et al. (2020) and Zhang et al. (2019) adopted a technique called focal loss, originally developed by Lin et al. (2017). This loss has two terms: the first up-weights the minority classes and the second focuses on difficult-to-classify samples. While Wang et al. (2020) applied the concept to lung nodules detection, Zhang et al. (2019) aimed at alleviating class imbalance in histopathology cancer image classification. However, focal loss was designed for uni-label classification and cannot be directly extended to deal with multi-label tasks. Yang et al. (2019) trained GANs to synthesize artificial training data and overcome the class imbalance challenge.

**SSL for 3D medical image analysis:** UA-MT (Yu et al., 2019) handles 3D MR images by using a volumetric convolutional network as the backbone segmentation network. FocalMix (Wang et al., 2020) claimed to be the first work to utilize SSL for 3D medical image detection. It uses a feature pyramid network to process 3D inputs. Multi-view learning techniques has also been adopted for 3D medical imaging, such as Deep Multi-Planar CoTraining (DMPCT) (Zhou et al., 2019) and Uncertainty-aware Multi-view Co-training (UMCT) (Xia et al., 2020a). All of these methods, however, rely on heavily-engineered backbone networks, which makes it difficult to handle 3D images with variable input size, e.g., MR images with varying number of slices. We also notice that most of the 3D adaptations have been designed for segmentation and detection, with less attention dedicated to classification tasks.

In summary, the above studies have made substantial progress towards leveraging semi-supervised learning for medical imaging needs. However, there remains an unmet demand for a generic and unified framework for radiology image classification. With the proposed NoTeacher framework, we offer a customizable framework to address the unique challenges for this application.

## 3. Methods

### 3.1. Problem Formulation

We consider input sample $x$, where $x$ can denote a single 2D image, a single slice in a 3D scan, or a 3D scan (an array of 2D slices). The classification target associated with $x$ is denoted as $y$, where $y$ can be uni-label (associated to only one label among many) or multi-label (associated to more than one label). For a given model training dataset, we have $L_T$ labeled samples $\{(x_i, y_i)\}_{i=1}^{L_T}$ and $U$ unlabeled samples $\{x_i\}_{i=1}^{U}$ with $U \gg L_T$. The semi-supervised learning task is to utilize the abundant pool of unlabeled samples to improve the generalizability of a model learned with the limited labeled samples.

### 3.2. Overview of the NoTeacher Framework

Figure 1 provides an overview of our proposed NoTeacher (NoT) framework. Given input sample $x$, we apply two random augmentations $\eta_1$ and $\eta_2$ to generate the augmented samples $x_1$ and $x_2$, respectively. These augmented samples serve as inputs to two networks $F_1$ and $F_2$ with similar architecture. The posterior output computed by $F_1$ is denoted as $f_1^L$ if $x$ is labeled and as $f_1^U$ if it is unlabeled. Similar naming conventions apply to $f_2^L$ and $f_2^U$. At the end of the forward pass, a loss function based on a graphical model is computed to encourage prediction consistency between two networks. Finally, the total loss is backpropagated to update the network parameters.

Importantly, the NoT framework in Figure 1 is highly flexible and can be customized to handle the diversity of unique challenges in radiology applications. In the upcoming subsections, we will introduce the NoT method (Subsection 3.3) followed by an adaptation for multi-label classification (Subsection 4.4.2) and an adaptation for 3D scans (Subsection 3.4). Afterward, we extend the NoT model with a generative assumption (termed as NoT-GA) to handle the scenario when the labeled and unlabeled sets have a distribution mismatch (Subsection 3.5). Finally, we provide insights on theoretical foundations and connections to established co-training methods (Subsection 3.6).

### 3.3. NoTeacher Loss

We describe the graphical model underlying the NoTeacher loss computation. For sample $x$, we recall that the networks $F_1, F_2$ take as inputs two augmented samples $x_1, x_2$ generated from two random augmentations $\eta_1, \eta_2$ (e.g., Gaussian noise, translation) respectively. Since $x_1, x_2$ are generated from the same sample, the network outputs $f_1 = F_1(x_1)$ and $f_2 = F_2(x_2)$ should be similar. Moreover, if a ground-truth target $y$ is available at training time, then those outputs should match the target. Because $y$ is binary, the network outputs can be interpreted as *posteriors* of the target, e.g., $f_1 = \Pr(y = 1|x_1)$. Inspired by previous works in kernel learning (Yu et al., 2011) and semi-supervised regression (Nguyen et al., 2019), we treat $y, f_1, f_2$ as random variables and design an undirected graphical model to model their probabilistic dependencies. Figure 2 (a) and (b) respectively illustrate the graphical models of NoT for labeled and unlabeled samples. The observed variables $y, f_1, f_2$ are represented by separate nodes, each is connected only to a latent variable called the *consensus function* $f_c \in [0, 1]$. As the name implies, $f_c$ enforces mutual agreement of the posteriors upon both labeled and unlabeled data. When the sample is labeled, $f_c$ acts as an information relay between the posteriors and $y$. For analytical tractability, we assume the differences $f_1 - f_c$ and $f_2 - f_c$ follow Gaussian distributions $\mathcal{N}(0, \sigma_1^2)$ and $\mathcal{N}(0, \sigma_2^2)$ respectively. To account for labeling noise, the difference $y - f_c$ is also assumed to follow a Gaussian distribution $\mathcal{N}(0, \sigma_y^2)$. Given a training batch of $n_L$ labeled and $n_U$ unlabeled samples, the likelihood of the NoT graphical models can be computed as follows:

$$p(\mathbf{y}|\mathbf{x}) \propto \exp\left(-\lambda_{y,1}^L \|\mathbf{f}_1^L - \mathbf{y}\|^2 - \lambda_{y,2}^L \|\mathbf{f}_2^L - \mathbf{y}\|^2\right) \\ \cdot \exp\left(-\lambda_{1,2}^L \|\mathbf{f}_1^L - \mathbf{f}_2^L\|^2 - \lambda_{1,2}^U \|\mathbf{f}_1^U - \mathbf{f}_2^U\|^2\right), \quad (1)$$

where $\mathbf{f}_\bullet^L, \mathbf{f}_\bullet^U$ are vectors containing the posteriors on labeled and unlabeled data respectively, $\mathbf{y}$ is the vector of target labels, and $\lambda_{y,1}^L, \lambda_{y,2}^L, \lambda_{1,2}^L, \lambda_{1,2}^U$ are derived as detailed in Supplement Subsection S1.1. Maximizing the likelihood in (1) yields the following loss function:

$$\mathfrak{L}_{sq} = \lambda_{y,1}^L \|\mathbf{f}_1^L - \mathbf{y}\|^2 + \lambda_{y,2}^L \|\mathbf{f}_2^L - \mathbf{y}\|^2 \\ + \lambda_{1,2}^L \|\mathbf{f}_1^L - \mathbf{f}_2^L\|^2 + \lambda_{1,2}^U \|\mathbf{f}_1^U - \mathbf{f}_2^U\|^2. \quad (2)$$

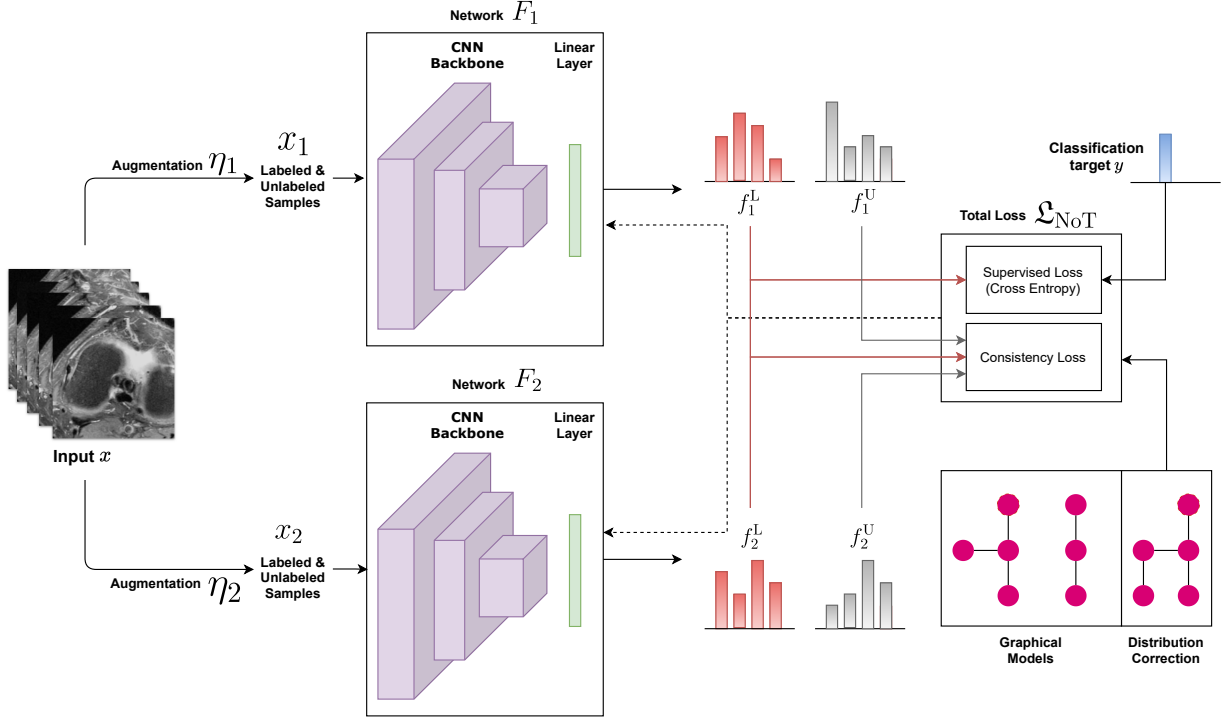

Figure 1: **Overview of NoTeacher**: Semi-supervised Learning Framework for Multi-dimensional Radiology Images. $x$ denotes a 2D or 3D sample. Augmented inputs $x_1$ and $x_2$ are fed into networks $F_1$ and $F_2$ with similar architecture. Solid arrows indicate forward pass: where flow of information from labeled inputs is in red, and flow of information from unlabeled inputs is in gray. The total loss function comprises a supervised cross-entropy loss and a consistency loss and is derived based on a graphical model. Dashed arrows denote the backward pass. The framework is flexible to accommodate different backbone architectures, classification task types, graphical models as needed.

To avoid vanishing gradients when using sigmoid activations with a squared loss, on the labeled data, we apply

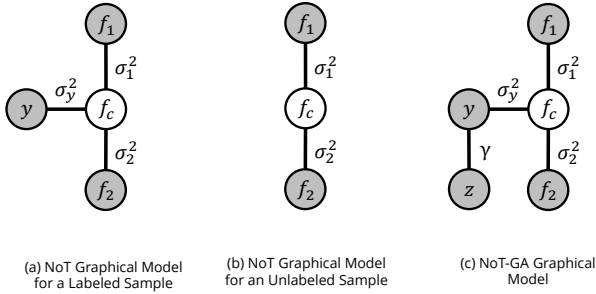

(a) NoT Graphical Model for a Labeled Sample
(b) NoT Graphical Model for an Unlabeled Sample
(c) NoT-GA Graphical Model

Figure 2: **Graphical Models for NoTeacher**: NoT graphical model on a single (a) labeled and (b) unlabeled sample. We use $y$ to denote the label; while $f_1, f_2$ represent the deep views and $f_c$ denotes the consensus function. Each edge represents an isotropic Gaussian potential with the corresponding variance. (c) The graphical model for NoT-GA with the additional variable $z$ to capture the class distribution mismatch information. If $z = 0$, the sample is drawn from the unlabeled set; if $z = 1$, the sample is drawn from the labeled set with label $y$. The edge between $z$ and $y$ represents a Bernoulli distribution where $\gamma_k = p(z = 1 | y = k)$.

CE loss on the posteriors instead. Our final NoT loss is therefore

$$\mathcal{L}_{\text{NoT}} = \lambda^L_{y,1} \text{CE}\left(\mathbf{y}, \mathbf{f}^L_1\right) + \lambda^L_{y,2} \text{CE}\left(\mathbf{y}, \mathbf{f}^L_2\right) \\ + \lambda^L_{1,2} \text{MSE}\left(\mathbf{f}^L_1, \mathbf{f}^L_2\right) + \lambda^U_{1,2} \frac{n_U}{n_L} \text{MSE}\left(\mathbf{f}^U_1, \mathbf{f}^U_2\right). \quad (3)$$

There are several important observations concerning the NoT loss. First, the first two terms of $\mathcal{L}_{\text{NoT}}$ represent the supervised losses while the last two terms enforce mutual agreement between the classifiers, thus enhancing consistency of the predictions. Second, the weights $\lambda$ are completely controlled by the hyperparameters $\{\sigma_1, \sigma_2, \sigma_y\}$, we provide the proof in Supplement Subsection S1.1. Last but not least, in Equation (2), the negative log likelihood includes $\ell_2$ norm terms which aggregate by sum. Meanwhile in Equation (3), the MSE loss terms aggregate by mean. Thus, we have introduced a normalization factor of $n_U/n_L$ in the final loss to reflect this change. In practice, we set $n_L = n_U$ in our implementation to cancel out this factor. To ensure fair comparison with supervised methods, during the training phase of NoT, each epoch covers one whole round of labeled samples, i.e., the model may not see the entire unlabeled training set. Hence, we can set $n_L = n_U$ without increasing the risk of overfitting on the labeled set.

It is worth noting that our proposed NoT method is similar to Mean Teacher (MT) (Tarvainen and Valpola, 2017) in the backbone design: NoT also employs two networks with similar architecture. However, compared with MT, we introduce two major changes: (a) the EMA update has been removed so that the networks are now completely detached and (b) the model is now trained with a novel loss function $\mathcal{L}_{\text{NoT}}$ based on a probabilistic graphical model to enhance consistency. Supplement Figure S1 highlights the similarities and differences between the training processes of NoT versus MT.

### 3.4. Adaptations for 3D Scan Classification

Our NoT framework allows free choice of backbones for the 3D scan classification task. While a standard choice may be to use a 3D convolutional neural network, it is also possible to employ a 2D network which applies max-pooling to condense feature vectors from multiple slices (e.g., as in Bien et al. (2018)). Where possible, using a 2D network is more efficient, allows one to leverage pre-training effectively, and offers greater flexibility for non-standard data formats.

We now describe how our NoT framework can be adapted for two unique characteristics of the 3D radiology scan classification task. First, 3D CT or MRI scans can have variable resolutions and variable slice depths (number of slices and slice thickness) for different patients. To allow variable-depth samples, we incorporate an indexed max-pooling layer that processes the slice depth dimensions. Second, unlike applying different random augmentations to a 2D image, data augmentation for 3D scans needs to be constrained at a volumetric level. Without this constraint, one might augment each slice differently and end up with a corrupted scan. Hence, for augmentations, we apply the same transformation on all the slices in a given scan.

### 3.5. NoTeacher Loss with Generative Assumption

In a general semi-supervised learning setup, the labeled set may have been constructed to have desired proportions of samples for each class, but the larger unlabeled set may not have the same proportions of samples per class as it would follow the prevailing class mix. Hence, the labeled set construction may result in class distributional mismatch between the labeled and unlabeled subsets of the training data. For example, in the case of chest X-ray classification, a rare pathology like Hernia may be intentionally well represented in the labeled set for optimal model training, but far less prevalent in a larger unlabeled set due to low disease prevalence. Hence, the labeled set construction may result in a mismatch in class distributions between the labeled and unlabeled subsets of the training data, which poses a challenge for semi-supervised methods (Chen et al., 2020). To address this distributional mismatch, we introduce NoTeacher with Generative Assumption (NoT-GA) by modifying the NoT graphical model to include a label-generating assumption.

Consider a uni-label, multi-class semi-supervised learning scenario with input $x$ and target $y$. Assume that there is a label-generating variable $z$ with the following rules: $z = 1$ indicates that the target is observed, i.e., the sample is labeled; while $z = 0$ indicates that the target is hidden, i.e., the sample is unlabeled. The label $y$ is a $K \times 1$ one-hot vector, where $K$ is the number of classes. We denote the class distributions in the labeled and unlabeled set by $\boldsymbol{\alpha}^L = [\alpha_1^L, \alpha_2^L, \ldots, \alpha_K^L]$ and $\boldsymbol{\alpha}^U = [\alpha_1^U, \alpha_2^U, \ldots, \alpha_K^U]$ respectively. $\boldsymbol{\alpha}^L$ and $\boldsymbol{\alpha}^U$ are probability distributions with nonnegative elements that sum up to 1, where $\alpha_k^L$ is the proportion of labeled samples that belong to the $k$-th class. We term the class distributions as matched if $\boldsymbol{\alpha}^L = \boldsymbol{\alpha}^U$, and the class distributions as mismatched otherwise. We note that the condition for distribution mismatch does not involve any condition on the relative proportions of classes, or class imbalance. In particular, class distributions can be matched in conditions with or without class imbalance: if both the labeled set and unlabeled set are class balanced (have uniform class distributions), or if both sets have identical non-uniform class distributions. If class imbalance exists in the unlabeled set while the labeled set is balanced (or vice versa), there will be distributional mismatch between the two training sets. In this case, the mismatch applies over the entire distribution, including both rare and common classes.

The NoT-GA graphical model takes the form in Figure 2(c). The probabilistic constraint between $z$ and $y$ is modeled as follows:

$$p(z = 1|y = k) = \frac{N_k^L}{N_k^L + N_k^U} = \frac{\alpha_k^L L_T}{\alpha_k^L L_T + \alpha_k^U U}$$
$$= \frac{1}{1 + \frac{\alpha_k^U U}{\alpha_k^L L_T}} = \gamma_k, \quad (4)$$

where $N_k^L$ and $N_k^U$ are the numbers of labeled and estimated unlabeled samples for class $k$, respectively. $L_T$ and $U$ are the number of training samples in the labeled and unlabeled set, respectively. One can either estimate $N_k^U$ from the class prevalence statistics or use a validation set that is randomly sampled from the unlabeled set. By definition, $\gamma_k$ represents the probability that a sample is drawn from the labeled set, given its ground-truth label $y = k$. In the scenario where there is distributional mismatch between labeled and unlabeled training sets, classes that are over-represented in the labeled set relative to the unlabeled set would be associated with higher $\gamma_k$ values than classes that are under-represented in the labeled set.

Under the generative assumption, the NoT-GA loss function is

$$\mathcal{L}_{\text{NoT-GA}} =$$
$$\lambda_{y,1}^L \text{CE}\left(\mathbf{y}, \mathbf{f}_1^L\right) + \lambda_{y,2}^L \text{CE}\left(\mathbf{y}, \mathbf{f}_2^L\right) + \lambda_{1,2}^L \text{MSE}\left(\mathbf{f}_1, \mathbf{f}_2\right)$$
$$- \log \left\{ \sum_{k=1}^{K} \exp\left[-\lambda_{y,1}^L \text{MSE}\left(\mathbf{f}_1^U, \mathbf{y}_k\right) - \lambda_{y,2}^L \text{MSE}\left(\mathbf{f}_2^U, \mathbf{y}_k\right)\right] (1 - \gamma_k) \right\}$$
(5)

where $\mathbf{y}_k$ denotes the concatenation of one-hot vectors with the $k$-th element equals to 1. We provide the detailed derivation in Supplement Subsection S1.2.

We make a few observations about NoT-GA. First, the $\gamma$ values are hyperparameters are estimated prior to training. Second, the first three terms of $\mathcal{L}_{\text{NoT-GA}}$ take after the terms in $\mathcal{L}_{\text{NoT}}$, so that the gradients generated from labeled samples remain the same. This allows NoT-GA to learn from the labeled samples as effectively as NoT. Third, the key difference for NoT-GA lies in the fourth term which is based on the unlabeled samples. In this term we integrate (average) over all possible values of the label. Essentially, we substitute $y = k$ for $k = 1, \ldots, K$ and compute the negative likelihood in each case. The model is encouraged to output posterior predictions that are closer to the distribution in the unlabeled set. Fourth, as NoT-GA adjusts its predictions according to $\boldsymbol{\alpha}^U$, we note that this model would perform optimally if $\boldsymbol{\alpha}^U$ reflects the class distribution in the validation set. This will be the case if the validation set is randomly sampled from the same distribution as the unlabeled data.

*3.6. Connections between NoTeacher and Co-training*

Here, we discuss how NoT shares theoretical foundations with established multi-view learning methods in the literature Xu et al. (2013). Multi-view learning focuses on representing the multiple distinct feature sets describing a given dataset. In this context, each feature set is termed as a *view*. We review the relations between NoT and two multi-view learning methods that are most related.

**Connection with co-training:** Co-training (Blum and Mitchell, 1998) is a primary multi-view learning technique which is mostly concerned with maximizing agreement between two views, often referred to as *consensus* or consistency enforcement. Like co-training, NoT employs two views of the data and maximizes the agreement between them. However, unlike conventional iterative co-training (as shown in Figure S2), which suffers from slow and unguaranteed convergence, NoT introduces an optimization where the loss function is defined based on the graphical model. This allows update of the NoT network parameters with back-propagation on every minibatch.

**Connection with Bayesian Co-training:** The graphical model adopted by our NoT method possesses similarities to graphical representations employed in Bayesian Co-training (BCT) (Yu et al., 2011), a variant of the classic co-training method. BCT introduced Gaussian process priors on multiple views, developed a graphical representation on top of the views, and constructed a transductive learning framework with a co-training kernel. Yet, NoT is distinct in that it employs an end-to-end deep learning framework instead.

**Connection to Multi-view Learning:** In general, multi-view learning techniques are concerned with not only maximizing agreement between two views, but also generating views with complementary features to capture diverse information from the data (Xu et al., 2013). NoT enforces this complementary principle by injecting different random augmentations to the networks during training.

Overall, NoT leverages the advantages of multi-view methods (especially co-training) while overcoming some of the challenges therein. As previously discussed, NoT also inherits its backbone design from MT, which is another (non co-training) consistency-based method, while avoiding some of the drawbacks of MT. In other words, NoT can harness the best of multi-view learning and MT. In our experiments, we empirically assess alignment with these theoretical foundations by examining the level of agreement between the two networks of NoT, and its influence to the performance of NoT.

## 4. Experiment Setup

*4.1. Datasets*

We considered datasets which (a) represent the 3 main modalities in radiology (X-Ray, CT, MRI), (b) cover the span of image dimensions encountered in practice (2D and 3D), (c) contain varying levels of ground truth annotation (image-level in 2D case, slice-level in 3D case, scan-level in 3D case), and (d) have a mix of uni-label and multi-label images. We also ensured all datasets chosen are publicly available for easy benchmarking and reproducibility. The 3 datasets that fulfilled these criteria are NIH-14 Chest X-Ray, RSNA Brain CT and Knee MRNet, as detailed below. Supplement Table S1 details the data splits for each of the datasets.

*4.1.1. NIH-14 Chest X-Ray (Wang et al., 2017)*

This dataset comprises 112,120 frontal chest X-Ray images. Of these, 46.1% are abnormal images, with abnormalities corresponding to one or more of 14 pathologies: Cardiomegaly, Emphysema, Edema, Hernia, Pneumothorax, Effusion, Mass, Fibrosis, Atelectasis, Consolidation, Pleural Thickening, Nodule, Pneumonia and Infiltration. Among abnormal images, 40.1% are multi-label, i.e., positive to more than one pathology. The remaining 53.9% of images are images with no findings i.e., negative for all 14 labels. This setup corresponds to a multi-label classification scenario with 14 binary labels. We used publicly available training (70%), validation (10%) and test (20%) splits with no patient overlaps (Zech, 2018).

*4.1.2. RSNA Brain CT (Flanders et al., 2020)*

This dataset provides 19,530 CT brain exams from the RSNA 2019 Challenge (Stage 1). Each slice in a CT scan is annotated with the presence or absence of 5 different types of intracranial hemorrhage, corresponding to 5 binary labels in a a multi-label classification task. The types of hemorrhage are Epidural, Intraventricular, Intraparenchymal, Subarachnoid and Subdural. In this dataset, 14.28% of the images are abnormal, i.e., positive to at least one type of bleed. Among which, 30.1% are labeled with multiple bleeds. The rest of the images (85.8%) are normal samples which are negative to all 5 labels. With this CT dataset, we are interested in slice-level classification and consider only one study for each patient. We derive random training (60%), validation (20%) and test (20%) splits with no patient overlaps. On a side note, our pre-processing steps include: (i) convert the raw pixel values to Hounsfield Units using the slope and rescale intercept from the original DICOM files, (ii) apply a window to restrict all pixel values to the width around the center as reported in the individual DICOM files, (iii) normalize the pixels to range [0, 255], and (iv) resize the image to $256 \times 256$. These pre-processing steps have been used by the leading solutions in the RSNA Challenge (Tao, 2019).

*4.1.3. Knee MRNet (Bien et al., 2018)*

This dataset, released by the Stanford University, consists of 1,370 knee MRI examinations. Each scan is annotated to indicate presence or absence of Anterior Cruciate Ligament (ACL) tears and Meniscal tears. In this dataset, 80.6% of the images are abnormal, i.e., positive for at least one type of tear. With this MRI dataset, we are interested in scan-level classification for the presence (or absence) of abnormality. While each scan consists of 3 physical views (axial, sagittal and coronal planes), we chose to provide the axial views for training. We used publicly available training (1,130 scans), and testing (120 scans) with no patient overlaps similar to Azcona et al. (2020). Of the 1,130 scans available for training, we randomly sampled 20% and designated those as the validation set.

*4.2. Design of Realistic Semi-Supervised Experiments*

To ensure that our experiments and evaluations are of practical relevance, we propose a realistic sampling approach that reflects the peculiar needs of real-world medical image annotation scenarios.

*4.2.1. Sampling the Labeled Set*

Figure 3 illustrates the sampling process adopted in our experiments. We consider a pool of $N$ unlabeled images in total for model development. With reference to our dataset, this pool of $N$ images corresponds to a combination of the training and validation sets as defined above. In practice, expert clinical raters would proceed to annotate the images in this pool systematically. Given a labeling budget of $L$, we randomly draw images from the pool $N$ and annotate them until the budget limit is reached. Note that this sampling step is completely random, as opposed to stratified sampling, where the $L$ images are chosen based on their ground-truth label. For lower budgets, there is a finite chance that random sampling may result in a set of $L$ images with incomplete coverage of the label space (i.e., some pathologies or abnormalities may not be represented in the $L$ images). Hence, we repeatedly sample and annotate from the unlabeled image pool until minimal numbers of labeled samples are obtained for each label. We denote the final number of labeled images as $L_F \geq L$. For subsequent increments in the labeling budget $L$, we maintain the existing labeled set and progressively add on images and annotate them. This sampling procedure is in contrast to common practice in the computer vision literature where the whole set of labeled samples is drawn again for each labeling budget.

*4.2.2. Data Splitting*

Once a labeled set of $L_F$ images is obtained, we further split them into $L_T$ images for the training set and $L_V$ images for the validation set, i.e., $L_F = L_T + L_V$. In a practical clinical setting, the majority of labeled data would be allocated to the training set, i.e., $L_T \gg L_V$ with $L_V$ being very small. For all experiments, we set a minimum viable size $L_{V,\min}$ for the validation set, so as to ensure that the stability of validation during model training. We use the remaining $N - L_F$ unlabeled images in the data pool as unlabeled training samples. As budget $L$ grows, the size of the unlabeled set decreases. In all experiments, we adopt a fixed-size held-out test set, which is not counted as part of the data pool available for model development.

Our realistic sampling and data process precludes the conveniences of stratified labeling, large validation sets, fixed unlabeled image budgets and balanced class distributions that are often encountered in standard SSL benchmarking papers (Oliver et al., 2018), and introduces additional challenges.

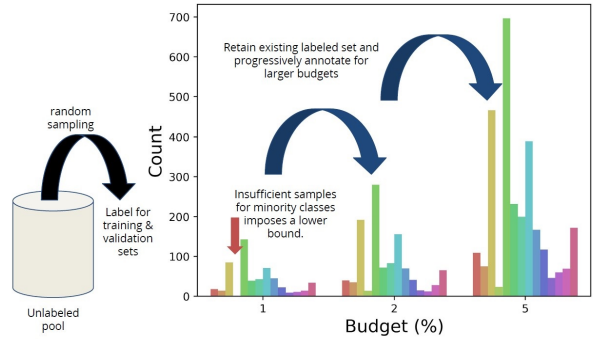

Figure 3: **Design of Realistic Sampling Process for Semi-supervised Experiments**

Table 1: DM-7511 Setup for NIH-14 Chest X-Ray

| Class | Train Set (Lab) | Train Set (Unlab) | Val Set | Test Set |
|---|---|---|---|---|
| No Finding | 243 | 1452 | 73 | 2835 |
| Infiltration | 243 | 1019 | 47 | 1843 |
| Pneumothorax | 243 | 214 | 6 | 484 |
| Mass | 243 | 231 | 11 | 423 |
| Total | 972 | 2916 | 137 | 5585 |

*4.3. Design of Class Distribution Mismatch Experiment*

In the general case, the process to construct a labeled subset of the training data could result in labeled and unlabeled subsets which do not have the same class distributions. To investigate performance in such scenarios, we design a class distribution mismatch experiment by constructing a multi-class uni-label classification dataset from the NIH-14 Chest X-Ray collection.

Specifically, we extract the subset of the NIH-14 Chest X-Ray dataset corresponding to 5% labeling budget (defined as per previous subsection). Within this subset, we pick samples associated with 4 classes: 2 each with high and low prevalence, respectively. For classes with low prevalence, we picked those that had sufficient numbers of samples to vary the class distributions. We characterized the dataset and observed that classes 'No Finding' and 'Infiltration' are the most prevalent, while classes 'Pneumothorax' and 'Mass' are amongst the rarer classes which also have sufficient numbers of samples. Further, for a clean analysis, we only consider the uni-label samples corresponding to one of these 4 classes.

To introduce the distribution mismatch, we constructed the labeled set to have balanced distribution of the 4 classes ($\boldsymbol{\alpha}^L \propto [1, 1, 1, 1]$), but allowed the unlabeled set to reflect the inherent imbalance amongst these classes in the NIH-14 Chest X-Ray dataset ($\boldsymbol{\alpha}^U \propto [7, 5, 1, 1]$ in alignment with the following order of classes: No Finding, Infiltration, Pneumothorax, Mass). Here, class distribution of the validation and test sets also follows $\boldsymbol{\alpha}^U$, since we assume that unlabeled set represents the general class distribution of unseen data. The resulting dataset is thus named "DM-7511" and is described in Table 1. In total, there are 972 labeled images, 2916 unlabeled images, 137 images for validation and 5585 images for the test set. Pneumothorax and Mass are over-represented in the labeled set (relative to the unlabeled set) and hence have higher $\gamma$ values, while Infiltration and No Finding are under-represented in the labeled set relative to the unlabeled set and are low-$\gamma$ classes.

### 4.4. Implementation Details

We now detail baseline methods and adaptations to cater to the specifics of our task, supervised training setup, hyperparameter tuning procedures, setup details of the semi-supervised experiments, and evaluation metrics.

#### 4.4.1. Baselines

We characterize the performance of NoT against 4 baselines including the supervised baseline (SUP) trained with the corresponding percentage of labeled samples and 3 semi-supervised methods that use the labeled samples along with the unlabeled samples (pseudolabeling (PSU), virtual adversarial training (VAT), mean teacher (MT)). In order to allow for a fair comparison with MT, we maintain EMA copies of SUP and VAT models, and report the best result from either the trained model or the EMA copy. This way, performance gains reported are not just due to averaging but also due to the ensembling mechanism. Regardless of the semi-supervised method, all experiments on any given dataset employed the same input data resolutions used in previous supervised learning works (Wang et al., 2017; Tao, 2019; Bien et al., 2018).

#### 4.4.2. Adaptations for Multi-Label Classification

In our datasets, a sample $x$ may be associated with more than one label, where each label presents a binary classification task. Therefore, we adapted the baselines for the multi-label classification scenario.

We adapted the supervised loss function from a uni-label cross-entropy formula to a multi-label binary cross-entropy formula. Our supervised cross-entropy loss (CE) per sample is

$$\text{CE}(y, f) = \sum_{k=1}^{K} \text{BCE}(y_k, f_{\{k\}}) \tag{6}$$

$$= -\frac{1}{K} \sum_{k=1}^{K} y_k \log f_{\{k\}} + (1 - y_k) \log (1 - f_{\{k\}}), \tag{7}$$

where $K$ is the number of labels, $y \in \{0, 1\}^K$ is a binary target label vector and $f \in [0, 1]^K$ is a vector of binary posterior probabilities. Note that $f_{\{k\}}$ denotes the $k^{\text{th}}$ label (different from earlier $f_1$ and $f_2$).

For VAT, we adopted a multi-label KL divergence for the LDS loss computation. This adaptation calculates KL divergence for each of the $K$ labels assuming a Bernoulli distribution, then averages the divergence across classes for LDS calculation. We empirically verified that VAT with our multi-label KL divergence offers better performance than other conventional losses in Table S4.

#### 4.4.3. Supervised Training Backbone

To guarantee a fair evaluation, the same backbone network architecture is used across all comparing methods, both supervised and semi-supervised. For NIH-14 Chest X-Ray, we use DenseNet121 following the implementation by Rajpurkar et al. (2017). For RSNA Brain CT, we use DenseNet169 as per the leading RSNA Challenge solution by Tao (2019). For Knee MRNet, we utilize an improved version of the MR-Net architecture with the AlexNet backbone used by Bien et al. (2018). In all cases, the networks are initialized with parameters pretrained on ImageNet. Prior to training, all input images are normalized based on ImageNet statistics. During training, runtime augmentations are applied, including random horizontal flipping, center cropping and resizing. We employ the Adam optimizer with learning rate $10^{-4}$, weight decay $10^{-5}$, $\beta = [0.9, 0.999]$ and $\epsilon = 10^{-8}$.

#### 4.4.4. Hyperparameter Tuning

For all semi-supervised methods, we tuned hyperparameters in conformity with literature norms to enable fair comparisons. In particular, we tuned EMA decay and consistency weight for MT, $\epsilon$ for VAT, and considered variations required for different labeling budgets. For NoT, although there are four weights $\lambda$, we only need to tune the three variances $\sigma$ that control them. From the loss function of NoT, one can see that only the ratio between the weights (and hence the ratio between the variances) is of consequence. Moreover, because the two networks have similar architecture, we can simply set $\sigma_1^2 = \sigma_2^2$ and vary the labeling noise $\sigma_y^2$. Supplement Table S2 enlists the hyperparameter tuning ranges and final choices for each of the experiments.

#### 4.4.5. 3D Experiments

For the 3D experiments on the Knee MRNet dataset, we adapt the original AlexNet architecture from Bien et al. (2018) as shown in Supplement Figure S2. Specifically, we focus on enhancing this architecture to efficiently meet the higher computational demands encountered in semi-supervised training. The original architecture processed only one volume at each step. In contrast, in the semi-supervised paradigm, a single pass has to consider both labeled and unlabeled samples at each step. Therefore, to make the architecture amenable to parallel training, we chose to increase the batch size from 1 to 16 by employing indexed max-pooling. Our adapted MRNet training pipeline is illustrated in Figure S3 in the Supplement. This change enables even faster training cycles and better model generalization.

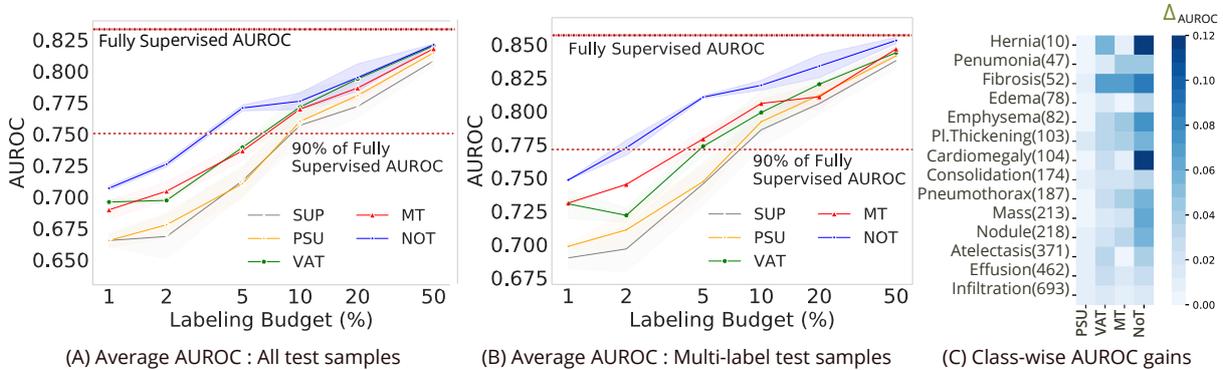

Figure 4: **Performance Results for the NIH-14 Chest X-Ray 2D-Image Classification Task**: (A) Average AUROC vs. Labeling Budget for the full test set. (B) Average AUROC vs. Labeling Budget performance for the multi-label portion of the test set. (A & B) The fully supervised model (dash-dotted line) is trained with 78,468 labeled images and evaluated on the associated test sets. NoT has largest performance gains when trained with the 5% labeling budget (3,923 labeled images). (C) Per-class AUROC gains for each SSL method over SUP baseline for the 5% labeling budget. Hernia (10) indicates that the labeled set for this budget has 10 images positive for class hernia.

### 4.4.6. Choice of Labeling Budgets

For each of the 3 datasets, we evaluate NoT and the baseline methods as a function of labeling budget. For the supervised model, we also report performance on the largest labeling budget that we can afford, i.e., on 100% budget (termed as 'fully supervised' learning). We determined the lowest labeling budget as follows. For the 2D image classification tasks (with labels at image or slice level), we start from $L = 500$ images or the smallest number of $L$ that allows at least one positive image per label, whichever is higher. For the NIH-14 Chest X-Ray dataset, the labeling budget is set at the image level. For the RSNA Brain CT dataset, the labeling budget is set at scan level and selected scans have all slices labeled. For the 3D image classification task (with labels at scan level), we start from $L = 32$ scans or 1082 slices. For the Knee MRNet dataset, the labeling budget is set at scan level and selected scans have only a scan-level annotation for the full volume. For all tasks, we incrementally increased labeling budget $L$ until the AUROC for semi-supervised methods approaches the AUROC of the fully supervised model.

### 4.4.7. Evaluation Metrics

For each of the 3 datasets, we report the classification performance of all methods across multiple labeling budgets. For multi-label tasks, we compute the per-label AUROC and report average over all labels. For the class distribution mismatch experiment, we compute the per-class AUPRC to assess performance given the imbalanced class distribution. All results are computed on fixed held-out test sets. In each case, we run three different seeds of random parameter initialization, and report the mean and standard deviations across seeds.

## 5. Results

We first present results from the 2D and 3D classification experiments. Then, we characterize performance of NoT-GA in relation to NoT and other baselines for the class distribution mismatch experiment.

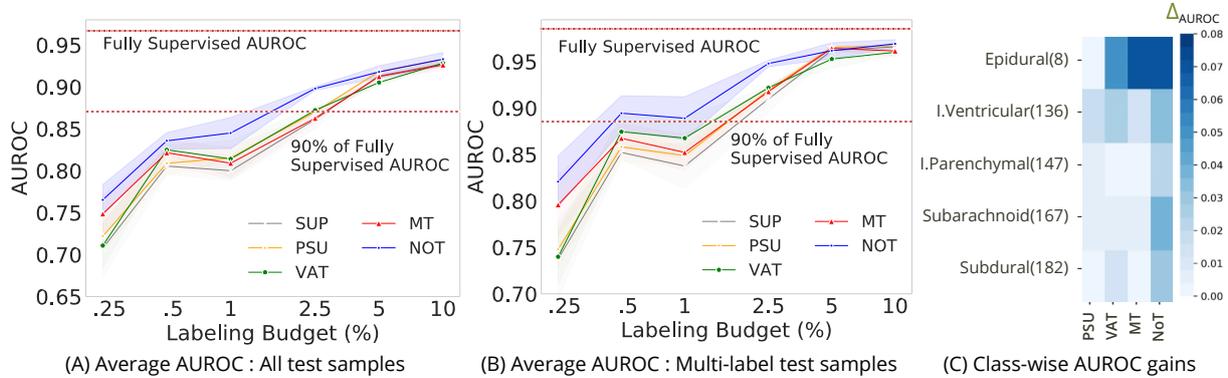

Figure 5: **Performance Results for the RSNA Brain CT 2D-Slice Classification Task**: (A) Average AUROC vs. Labeling Budget for the full test set. (B) Average AUROC vs. Labeling Budget performance for the multi-label portion of the test set. (A & B) The fully supervised model (dash-dotted line) is trained with 352,839 labeled slices (10,247 scans) and evaluated on the associated test sets. NoT has largest performance gains when trained with the 1% labeling budget (3,495 slices from 102 scans in labeled set). (C) Per-class AUROC gains for each SSL method over SUP baseline for the 1% labeling budget. Epidural (8) indicates that the labeled set for this budget has 8 images positive for class epidural bleed.

### 5.1. 2D Image Classification Results

Figures 4 and 5 show results on the NIH-14 Chest X-Ray and RSNA Brain CT datasets respectively. Detailed performance numbers are provided in Supplement Table S3.

Figure 4(A) and Figure 5(A) show the results on the full NIH-14 Chest X-ray and RSNA Brain CT test sets. In the low labeling budget regimes, the semi-supervised methods outperform the supervised baseline (SUP). In the higher labeling budget regimes, the performance of all semi-supervised methods converges, suggesting that the gain from unlabeled data saturates when large numbers of labeled samples are available.

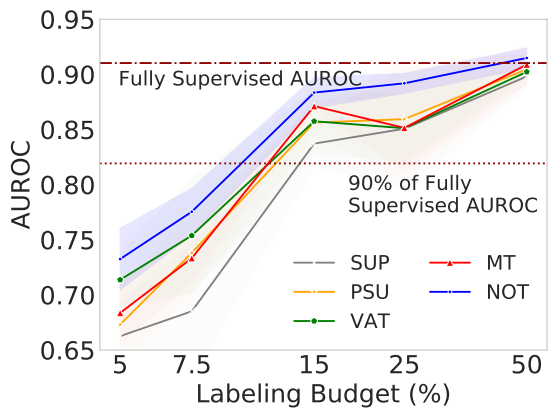

Figure 6: **Performance Results for Knee MRNet 3D Scan Classification Task**: Average AUROC vs. Labeling Budget for the full test set. Fully supervised model (dash-dotted line) is trained with 904 labeled scans and evaluated on the same test set. NoT has largest performance gains when trained with the 25% labeling budget (227 labeled scans).

We evaluate performance of NoT in relation to (a) the fully supervised method and (b) to other semi-supervised methods. First, we note that NoT can surpass 90% of the fully supervised AUROC with less than 5% labeling budget for NIH-14 Chest X-Ray dataset and less than 2.5% labeling budget for RSNA Brain CT dataset. Second, we note that NoT method outperforms the other semi-supervised methods. For NIH-14 Chest X-Ray, with 5% labeling budget (3,923 labeled images), NoT gains 6.4% over the corresponding supervised baseline (SUP) and over 3.1% vs. other SSL methods. For RSNA Brain CT, with 1% labeling budget (3,495 labeled slices, 102 scans), NoT gains 4.5% over the supervised baseline (SUP) and over 3.0% vs. other SSL methods.

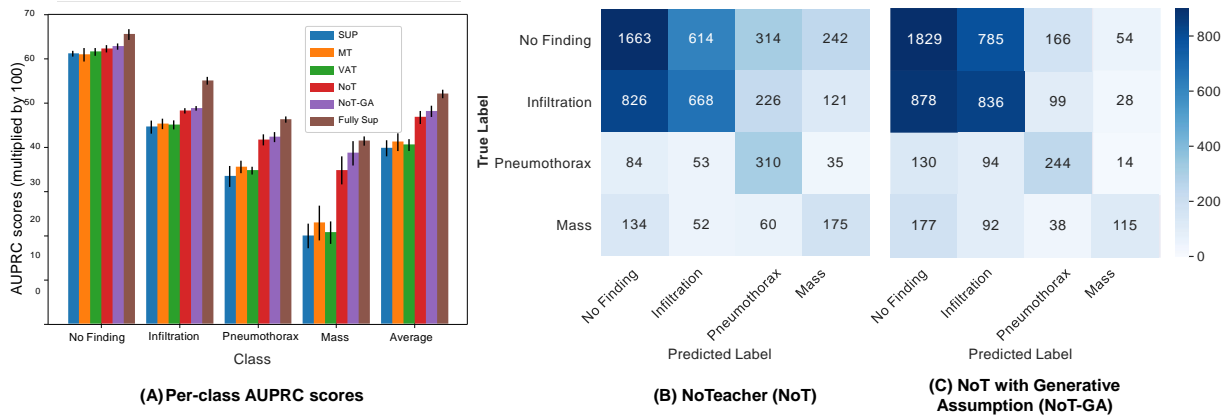

Figure 7: **Results on DM-7511 setup:** (A) Average AUPRC per class over 5 seeds; average confusion matrix (threshold 0.5) over 5 seeds with vanilla NoT (B) and with NoT-GA (C).

Next, we zoom in on performance for the subset of multi-label cases in each of the NIH-14 Chest X-ray and RSNA Brain CT test sets. The results are in Figure 4(B) and Figure 5(B). We observe that these plots show similar trends as those seen with the full test sets comprising both uni-label and multi-label cases. This suggests that NoT also has substantial gains for multi-label classification.

We then consider the per-class AUROC gains for each SSL method over the corresponding supervised baseline (SUP). We evaluate the per-class AUROC gain at 5% budget for NIH-14 Chest X-Ray and 1% budget for RSNA Brain CT. Figure 4(C) and Figure 5(C) show the results in a heatmap format. In each case, the classes (conditions) are ordered by the associated number of images in the labeled sets. In general, we observe that NoT gains more than other SSL methods. In particular, NoT gains much more than the baselines for rarer conditions with lower prevalence in the labeled subsets.

Finally, in the Supplement Table S3, we also enlist the performance numbers for other closely related methods (Aviles-Rivero et al., 2019; Liu et al., 2020) for comparisons.

### 5.2. 3D Scan Classification Results

We characterized performance of NoT and the baselines on the 3D Knee MRNet dataset. Result are in Figure 6 and detailed performance numbers are in Supplement Table S3. First, we note that NoT can surpass 95% of the fully supervised AUROC with less than 15% labeling budget (just 137 labeled scans). Second, NoT has the highest performance across all labeling budgets, followed by VAT, MT and PSU. In particular, with 7.5% labeling budget (just 55 labeled scans), NoT gains 9.0% over the corresponding supervised baseline (SUP) and over 2.2% vs. other SSL methods. Further, we note that the performance of NoT is less variable (across seeds) than the next best SSL method VAT.

### 5.3. Class Distribution Mismatch Results

Finally, we present results for the class distribution mismatch experiment described above. We characterize performance of NoT-GA in relation to the vanilla NoT, the leading SSL methods (VAT, MT), the supervised baseline (SUP) trained on labeled data only, and the fully supervised baseline (Fully SUP) trained on the fully annotated dataset where all training samples are annotated. The per-class AUPRC results are in Figure 7, Panel A. The corresponding results in tabular form are available in Table S5. The confusion matrices for NoTeacher based methods (threshold 0.5, averaged across seeds) are in Figure 7, Panels B-C. Both NoT and NoT-GA outperform other SSL methods by significant margins (5%-7% averaging across all classes). In particular, NoT-GA gains much more over other methods (including NoT) for the rarer classes Pneumothorax and Mass. This suggests that the adaptations introduced in NoT-GA can effectively handle mismatches in distribution between labeled and unlabeled datasets. Further, the confusion matrix results show that NoT-GA has greater ability to adapt to the distributional mismatch than NoT. For example, while NoT tends to overpredict on Pneumothorax and Mass (the two right-most columns of Panel B), NoT-GA assigns this label to a smaller number of samples, but with higher precision (the two right-most columns of Panel C).

## 6. Additional Analyses

In this section, we report additional analyses to assess the influence of training parameters such as the labeled:unlabeled ratio in a minibatch and degree of augmentation on the NoT performance. Moreover, to gain intuition for the value of the generative assumption in NoT-GA, we study NoT and NoT-GA in a wider range of distribution mismatch settings. For the above experiments, we choose the NIH-14 Chest X-Ray dataset with a 5% labeling budget. However, we expect that the trends are not dataset dependent. Finally, we analyse performance gains over Mean Teacher to highlight advantages of the co-training approach employed by NoTeacher.

### 6.1. Effect of Varying Annotation Ratio within a Minibatch

To assess the sensitivity of model performance to the ratio of unlabeled to labeled samples ($n_U/n_L$) in a training minibatch, we vary this ratio within a range and report the corresponding performance of NoT. Our experiment setup and results are shown in Table 2. The empirical results show that NoT is not affected by the increase (or decrease) in the annotation ratio within a minibatch. We note that in Equation (3), because the $\lambda$ values are computed from the $\sigma$ hyperparameters, they are not independent variables. Hence, the ratio $n_U/n_L$ cannot be absorbed into the $\lambda$ values.

### 6.2. Effect of Varying Augmentation Levels

We examine the effect of training with increasing amounts of data augmentation on SUP, PSU, VAT, MT and NoT performance. We consider 4 levels of augmentation. For the various levels of augmentation, we consider variations encountered during the acquisition of X-rays in the real world. Typical variations are due to (i) affine transformations like shifts and rotations; and (ii) brightness and contrast variations occurring as a result of windowing. We start with no augmentation, resize the input image to 224 × 224 and normalize to the standard normal distribution. Next, for a first level of augmentation we add random horizontal flipping, following the implementation in Zech (2018). For a second level of augmentation, we continue to add affine transformations such as rotation, translation and scaling. Finally, for the highest level of augmentation, we apply brightness, contrast and saturation variations (collectively termed as intensity variations) on top of the previous augmentations.

Table 2: Effect of ratio of unlabeled data to labeled data in the batch on final performance of the NoTeacher model. Average AUROC scores on the NIH-14 dataset with 5% annotation.

| $n_U$ | $n_L$ | $n_U/n_L$ | NoT |
|---|---|---|---|
| 8 | 16 | 0.5 | 77.24 ± 1.19 |
| 16 | 16 | 1 | 77.04 ± 0.22 |
| 32 | 16 | 2 | 77.32 ± 0.23 |
| 64 | 16 | 4 | 77.32 ± 1.05 |

The results are presented in Table 3. At each level of augmentation, NoT outperforms all the baselines. All methods show performance improvement with increasing augmentation levels with the exception of the highest augmentation level (where the effect of augmentation saturated). When no augmentations are applied, we notice that MT suffers in performance and shows negligible gains with respect to the supervised method. VAT shows better performance, possibly because the method internally generates perturbations based on the data while calculating the local distributional smoothness loss. At the higher augmentation levels, we note that the MT and VAT perform comparably to the SUP baseline, while NoT outperforms all baselines by a large margin. Thus, in settings with heavy data augmentation, the combination of consistency enforcement with the complementary principle in NoT may be essential to effectively leverage unlabeled data and boost performance.

### 6.3. Analysis of NoTeacher versus NoTeacher with Generative Assumption

To understand the value-add of NoT-GA in a setting with severe distributional mismatch but mild class imbalance, we construct a different dataset "DM-3311", de-scribed in Table 4. Here, each class has the same number of samples in the overall training set, but the mismatch between labeled and unlabeled distributions is much more significant. Specifically, the class imbalance ratio for labeled data $\boldsymbol{\alpha}^L$ is $\propto [1, 1, 3, 3]$), while the class imbalance ratio for unlabeled data $\boldsymbol{\alpha}^U$ is $\propto [3, 3, 1, 1]$ with the following class order: No Finding, Infiltration, Pneumothorax, Mass. We note that the validation and test sets also follows $\boldsymbol{\alpha}^U$, due to our assumption that unlabeled set represents the general class distribution of unseen data.

Table 3: Effect of level of augmentation on the methods. Demonstrated on the NIH-14 dataset with 5% annotation, reporting average AUROC. Each row also includes augmentations from all previous rows.

| Aug Level | SUP | PSU | VAT | MT | NoT |
|---|---|---|---|---|---|
| **No Augmentation** | 70.45 ± 0.93 | 70.06 ± 0.87 | 73.04 ± 0.45 | 71.07 ± 1.39 | **75.79 ± 0.29** |
| **+ Horizontal Flip** | 70.68 ± 1.31 | 70.93 ± 0.99 | 73.94 ± 0.14 | 73.60 ± 0.70 | **77.04 ± 0.22** |
| **+ Affine Transformation** | 74.46 ± 0.75 | 75.60 ± 0.45 | 75.72 ± 0.76 | 75.61 ± 0.82 | **78.37 ± 0.82** |
| **+ Intensity Variation** | 74.19 ± 0.30 | 75.07 ± 0.96 | 75.64 ± 0.69 | 75.48 ± 0.68 | **78.87 ± 0.06** |

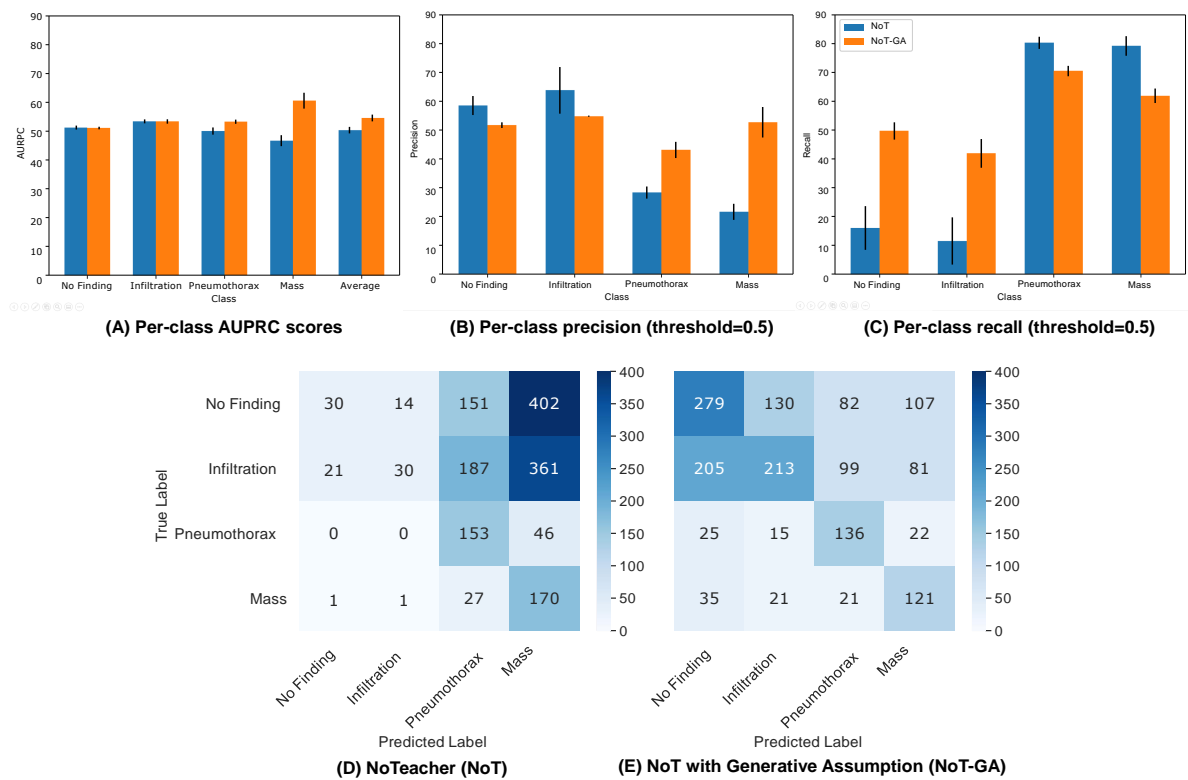

Figure 8: **Results on DM-3311 setup:** (A) Average AUPRC per class; (B) Average precision (threshold 0.5); (C) Average recall (threshold 0.5); Average confusion matrix (threshold 0.5) with vanilla NoT (D) and with NoT-GA (E). The results are computed over 5 seeds.

For the DM-3311 setup, we compared NoT-GA against vanilla NoT, and report the average AUPRC, precision and recall across 5 random seeds in Table S6. We observe that NoT-GA substantially improves the precision scores on Pneumothorax and Mass (the high-$\gamma$ classes). At the same time, it also improves the recall on No Finding and Infiltration (the low-$\gamma$ classes). We also plot the average confusion matrices in Figure 8. The confusion matrix for NoT suggests that it is biased to the class distribution of labeled training data and therefore overpredicts high-$\gamma$ classes (the two right-most columns), but underpredicts low-$\gamma$ classes (the two left-most columns). Meanwhile, NoT-GA is able to adjust to the unlabeled distribution by focusing more on the low-$\gamma$ classes. The side effects of this adjustment include a slightly lower precision on low-$\gamma$ classes and a lower recall on high-$\gamma$ classes. Overall, the average AUPRC of NoT-GA surpasses that of NoT by 8.4%. Comparing to the results in Figure 7, the effects of NoT-GA have been amplified in this experiment. We have also conducted this experiment on other variants of the DM-3311 setup in Section S3 of the Supplement.

Table 4: DM-3311 Setup for NIH-14 Chest X-Ray

| Class | Train Set (Lab) | Train Set (Unlab) | Val Set | Test Set |
|---|---|---|---|---|
| No Finding | 200 | 600 | 60 | 600 |
| Infiltration | 200 | 600 | 60 | 600 |
| Pneumothorax | 600 | 200 | 20 | 200 |
| Mass | 600 | 200 | 20 | 200 |
| Total | 1600 | 1600 | 160 | 1600 |

*6.4. Analysis of Training Dynamics: NoTeacher versus Mean Teacher*

We now conduct a comparative analysis between NoT and MT to understand some differences in their training and performance. For each of our 2D and 3D datasets, we train both NoT and MT using the same semi-supervised setup with the same labeled/unlabeled/validation data split, backbone architecture, and optimizer. During training, we create a validation checkpoint every $I$ iterations ($I$ varies depending on the dataset). At each checkpoint, we evaluate the AUROC scores on the validation set for each of the following four models: the two models $F_1$ and $F_2$ of NoT, the student model $F_S$ and the teacher model $F_T$ of MT. Further, we apply a binarization threshold $\tau$ on the posterior predictions and compute two *disagreement statistics*: NoT-Dis that counts the number of validation samples for which $F_1$ and $F_2$ produce unmatched predictions; and MT-Dis that counts the number of validation samples for which $F_S$ and $F_T$ of MT produce unmatched predictions. Since our task is multi-label, two sets of binary predictions are considered as unmatched when they differ on at least one label. Finally, we plot and compare the validation AUROC scores and disagreement statistics of NoT and MT.

Figure 9 shows the analysis results on three datasets: (a) NIH-14 Chest X-Ray, (b) RSNA Brain CT and (c) Knee MRNet. Overall, we have two observations. First, towards the end of training, we observe that the NoT models $F_1$ and $F_2$ achieve higher validation AUROC than the MT models $F_S$ and $F_T$. Second, the disagreement counts of NoT is consistently lower than that of MT prior to convergence (especially during the first 2000 iterations). To dig deeper, we considered the training process more closely. We note that, during the training of MT, the periods with high disagreement counts are associated with significant drops in validation AUROC. On the other hand, when the NoT model is training, the changes in disagreement count do not seem to have a strong correlation with changes in its validation AUROC. The above patterns are consistent over multiple runs (see Supplement Figures S6 and S7). These findings suggest that for MT and NoT, the disagreement statistics may relate to the dynamics of the training process in different ways. Possible explanations include: (i) the two methods are different in their optimization strategies, MT uses an EMA to update its teacher network, while NoT applies backpropagation on both views; (ii) the MT loss function is based on a consistency assumption, while the NoT loss function is derived from a probabilistic graphical assumption. Future developments focused on resolving these possibilities could help evolve hybrids or variants of these approaches. We highlight that by minimizing the disagree-ment count, NoT is maximizing the consensus between its views, thus conforming to the consensus principle of multi-view learning.

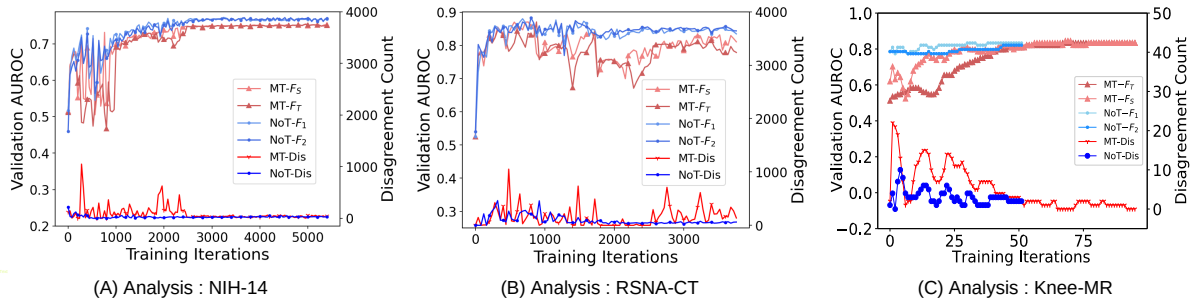

(A) Analysis : NIH-14    (B) Analysis : RSNA-CT    (C) Analysis : Knee-MR

Figure 9: **Performance analysis of NoT and MT**: Performance as a function of training as indicated by AUROC scores on the validation set (left vertical axis) and disagreement statistics (right vertical axis). (A) NIH-14 Chest X-Ray with 5% labeling budget, $X = 40$, $\tau = 0.25$, (B) RSNA Brain CT with 2.5% labeling budget, $X = 40$, $\tau = 0.2$ and (C) Knee MRNet with 7.5% labeling budget, $X = 1$, $\tau = 0.1$. All results are based on one validation checkpoint every $I$ training iterations.

## 7. Discussion and Conclusion

We have developed and demonstrated a novel semi-supervised learning framework, NoTeacher (NoT) for radiology image classification. We rigorously characterized performance on benchmark datasets representing the key modalities of radiology (X-Ray, CT and MRI) (Rajpurkar et al., 2017; Flanders et al., 2020; Bien et al., 2018). Across datasets, our results show that NoT can achieve over 90-95% of the fully supervised AUROC with just a few 1000 ( 3500) labeled images or slices (2D), or just over a 100 labeled scans (3D). This constitutes a stark reduction in the annotation burden as the fully supervised case would involve labeling at the scale of over 100,000 images or slices (2D) or 1000 scans (3D). In particular, for cross-sectional imaging, our approach can perform well with scan-level labels and hence does not require more intensive labeling of individual slices[2]. As such, our results suggest feasibility of deep learning with minimal supervision for radiology applications.

---
[2]For the RSNA CT experiments, we could also perform scan-level classification by using the relevant adaptations to the backbone architecture.

Further, the extensions of the NoTeacher framework for SSL scenarios with distribution mismatch make our approach suitable for low labeling budget cases where it may not be possible to match the distributions of labeled and unlabeled datasets. Although we have demonstrated the NoT-GA extension for uni-label scenarios, extension to the multi-label case is possible with one of two approaches: (i) by modifying the graphical model and re-deriving the losses accordingly, or (ii) by reducing multiple labels to a single label with multiple classes by assigning each unique combination of label assignments in the multi-label setup to a distinct class in the multi-class setup. In option (ii), the integration in the last term of $\mathcal{L}_{\text{NoT–GA}}$ requires iterating over all possible values of the (multi-class) label $y$. Exploring how best to extend NoT-GA to the multi-label setting would be an interesting direction for future work.

Our work has implications for streamlining the clinical image annotation process for rapid model development and iteration. As our experiments are set up within a realistic practical annotation process, the contributions can be more easily translated to impact realistic annotation workflows. Further, as our framework can flexibly cater to a range of classification tasks spanning 2D and 3D, single and multiple labels, and varying levels of annotation (at scan/slice/image level), and mismatch between the labeled and unlabeled sets, it could be embedded within available general-purpose radiology annotation platforms.

From a methods perspective, NoT is principled and has connections with established machine learning approaches. While VAT, PSU and even MT are essentially single-network models (the MT teacher network is learned passively via EMA), NoT is a multi-view learning technique which benefits from having multiple views of the data. We also note that co-training is known to be closely related to label propagation. Specifically, the iterative co-training algorithm can be positioned as an integrative label propagation process over the two views Wang and Zhou (2010). Thus, NoT is related, albeit subtly, to other label propagation techniques that have been adopted for medical imaging (e.g., Aviles-Rivero et al. (2019), Heckemann et al. (2006)).

Co-training frameworks have been shown to be applicable to wider task types beyond classification. For example, Xia et al. (2020b) demonstrated how their Uncertainty-Aware Multi-View Co-Training (UMCT) method can be applied for either classification or segmentation, by making minor changes to the supervised loss and backbone network. As NoT is a co-training framework, it has potential for expansions to other task types such as segmentation or detection. It is also possible to explore expansions of NoT to the multi-task semi-supervised learning paradigm to cater to joint classification and segmentation objectives, akin to Terzopoulos et al. (2019). While we showed that NoT can be applied to 3D data by learning from augmentations of the same physical view, expansions to multi-planar learning (as in Deep Multi-Planar CoTraining (DMPCT) (Zhou et al., 2019)) can also be envisioned. Future directions will explore some of these expansions and develop the NoT framework for semi-supervised learning on some of these relevant radiology applications.

## Acknowledgments

Research efforts were supported by funding and infrastructure for deep learning and medical imaging R&D from the Institute for Infocomm Research, Science and Engineering Research Council, A*STAR AME Programmatic Funds (Grant No. A20H6b0151), and by the Singapore International Graduate Award (SINGA Award to Shafa Balaram), Agency for Science, Technology and Research (A*STAR), Singapore. We acknowledge insightful discussions with Jayashree Kalpathy-Cramer and Praveer Singh at the Massachusetts General Hospital, Harvard Medical School, Boston USA. We also thank Ashraf Kassim from the National University of Singapore for his support.

**Supplementary Material**

*S1. NoTeacher Method: Additional Details/Theory*

*S1.1. Derivation of NoT weights*

In this appendix we derive the marginal density distribution of the network outputs when the consensus function is integrated out. Subsequently, the loss multipliers $\{\lambda^L_{y,1}, \lambda^L_{y,2}, \lambda^L_{1,2}, \lambda^U_{1,2}\}$ will be derived as functions of the hyperparameters $\{\sigma_1^2, \sigma_2^2, \sigma_y^2\}$. Given the NoT graphical model, it is necessary to integrate $f_c$ out of the joint density distribution of the graph, because $f_c$ is a latent variable. Consider a more general model, where there are $M$ networks, each outputs a posterior, i.e., $\{f_m\}_{m=1}^M$. Graphically, each posterior is represented by a random variable connected only to the consensus function $f_c$ via a zero-mean Gaussian potential

$$f_m - f_c \sim \mathcal{N}\left(0, \sigma_m^2\right). \tag{8}$$

Technically, even the target $y$ can be considered as a posterior with potential $y - f_c \sim \mathcal{N}\left(0, \sigma_y^2\right)$. The joint density distribution function of the general graph is as follows

$$p(f_c, f_1, \ldots, f_M) = \frac{1}{\mathcal{Z}_1} \prod_{m=1}^M \exp\left[-\frac{(f_c - f_m)^2}{2\sigma_m^2}\right] \tag{9}$$

$$= \frac{1}{\mathcal{Z}_1} \exp\left(-\frac{\psi}{2} f_c^2 + \phi f_c + \chi\right), \tag{10}$$

where the normalizing factor $\mathcal{Z}_1$ is a constant w.r.t. $f_c, f_1, \ldots, f_M$ and

$$\psi = \sum_{m=1}^M \frac{1}{\sigma_m^2} \quad \phi = \sum_{m=1}^M \frac{f_m}{\sigma_m^2} \quad \chi = \sum_{m=1}^M -\frac{f_m^2}{2\sigma_m^2}. \tag{11}$$

Notice that $\psi, \phi, \chi$ are constants w.r.t. $f_c$. In addition, we have the following integration rule

$$\int \exp\left(-\frac{1}{2} a x^2 + b x\right) dx = \sqrt{\frac{2\pi}{a}} \exp\left(\frac{b^2}{2a}\right), \tag{12}$$

where $a$ is positive. With this rule, knowing that $\psi > 0$, we can integrate $f_c$ out of the joint distribution in (10) to obtain the following marginal likelihood as follows

$$p(f_1, \ldots, f_M) = \int p(f_c, f_1, \ldots, f_M) df_c \tag{13}$$

$$= \frac{1}{\mathcal{Z}_2} \exp\left(\frac{\phi^2}{2\psi} + \chi\right) = \frac{1}{\mathcal{Z}_2} \exp\left[\frac{1}{2\psi}\left(\phi^2 + 2\psi\chi\right)\right] \tag{14}$$

$$= \frac{1}{\mathcal{Z}_2} \exp\left[\frac{1}{2\psi}\left(\sum_m \frac{f_m^2}{\sigma_m^4} + 2\sum_m \sum_{k>m} \frac{f_m f_k}{\sigma_m^2 \sigma_k^2}\right.\right.$$
$$\left.\left. - \psi \sum_m \frac{f_m^2}{\sigma_m^2}\right)\right] \tag{15}$$

$$= \frac{1}{\mathcal{Z}_2} \exp\left[\frac{1}{2\psi}\left(\sum_m \sum_{k>m} -\frac{f_m^2 - 2 f_m f_k + f_k^2}{\sigma_m^2 \sigma_k^2}\right)\right] \tag{16}$$

$$= \frac{1}{\mathcal{Z}_2} \exp\left[\sum_m \sum_{k>m} -\lambda_{m,k} (f_m - f_k)^2\right], \tag{17}$$

where $\mathcal{Z}_2$ is another constant w.r.t. $f_c, f_1, \ldots, f_M$, and

$$\lambda_{m,k} = \left[2\sigma_m^2 \sigma_k^2 \left(\sum_{i=1}^M \frac{1}{\sigma_i^2}\right)\right]^{-1}. \tag{18}$$

This result implies that the marginal likelihood can be factorized as a product of $\binom{M}{2}$ components, each component is a Gaussian distribution on the difference between a pair of posteriors $(f_m, f_k)$ with zero mean and variance $\lambda_{m,k}^{-1}$.

The NoT graphical models are special cases of this general model. For a labeled sample, we have $M = 3$, i.e., there are three observed variables $f_1, f_2$ and $y$. Thus, we use $\{\lambda^L_{y,1}, \lambda^L_{y,2}, \lambda^L_{1,2}\}$ as shorthand notations to denote $\{\lambda_{f_1,y}, \lambda_{f_2,y}, \lambda_{f_1,f_2}\}$ respectively. By applying (18), they can be expressed in terms of the hyperparameters $\{\sigma_1^2, \sigma_2^2, \sigma_y^2\}$ as

$$\lambda^L_{y,1} = \lambda_{f_1,y} = \frac{\sigma_2^2}{2\left(\sigma_1^2 \sigma_2^2 + \sigma_2^2 \sigma_y^2 + \sigma_1^2 \sigma_y^2\right)} \tag{19}$$

$$\lambda^L_{y,2} = \lambda_{f_2,y} = \frac{\sigma_1^2}{2\left(\sigma_1^2 \sigma_2^2 + \sigma_2^2 \sigma_y^2 + \sigma_1^2 \sigma_y^2\right)} \tag{20}$$

$$\lambda^L_{1,2} = \lambda_{f_1,f_2} = \frac{\sigma_y^2}{2\left(\sigma_1^2 \sigma_2^2 + \sigma_2^2 \sigma_y^2 + \sigma_1^2 \sigma_y^2\right)}. \tag{21}$$

Similarly, for an unlabeled sample, there are $M = 2$ observed variables $f_1$ and $f_2$, the formula of $\lambda^U_{1,2}$ is therefore

$$\lambda^U_{1,2} = \lambda_{f_1,f_2} = \frac{1}{2\left(\sigma_1^2 + \sigma_2^2\right)}. \tag{22}$$

## S1.2. Derivation of NoT-GA loss function

The NoT-GA graphical model can be divided into two cases: (i) a labeled sample when $z = 1$ and (ii) an unlabeled sample when $z = 0$. Thus, we can compute the likelihood separately for labeled and unlabeled data. The joint distribution of a sample is

$$p(z,y,f_c,f_1,f_2) = \begin{cases} \gamma_y p(f_c,y,f_1,f_2), & \text{if } z = 1 \\ (1-\gamma_y) p(f_c,y,f_1,f_2), & \text{if } z = 0 \end{cases}, \quad (23)$$

where we use $\gamma_y$ to denote the $\gamma_k$ value corresponding to the target label, i.e., $y = k$. By integrating out $f_c$ – the latent consensus function, we obtain the data likelihood as

$$p(z,y,f_1,f_2) \propto \begin{cases} \gamma_y \int p(f_c,y,f_1,f_2)\,df_c, & \text{if } z = 1 \\ \iint (1-\gamma_y) p(f_c,y,f_1,f_2)\,df_c\,dy, & \text{if } z = 0 \end{cases}. \quad (24)$$

Note that in (24), the ground-truth label is unobserved for the case $z = 0$, thus an additional integration over $y$ is required. By taking the integration over $f_c$ as performed in Subsection S1.1, the log likelihood can be computed. For a labeled sample, it is

$$\log[p(z=1,y,f_1,f_2)] \propto -\lambda_{y,1}^L \|f_1 - y\|^2 - \lambda_{y,2}^L \|f_2 - y\|^2 - \lambda_{1,2}^L \|f_1 - f_2\|^2 + \log(\gamma_y). \quad (25)$$

For an unlabeled sample, it is

$$\log[p(z=0,y,f_1,f_2)] \propto -\lambda_{1,2}^L \|f_1 - f_2\|^2 + \log\left[\sum_y \exp\left(-\lambda_{y,1}^L \|f_1 - y\|^2 - \lambda_{y,2}^L \|f_2 - y\|^2\right)(1-\gamma_y)\right]. \quad (26)$$

Notice that in equation (25), the last term is a constant w.r.t. the observed variables and can be removed from the optimization. By combining the negative log likelihood functions over all labeled and unlabeled samples in the dataset, we obtain the NoT-GA loss function in (5).

## S1.3. Connections to Other Methods

The training process of MT and NoT are compared in Figure S1, with the networks in NoT no longer being connected by the EMA update. Mean Teacher (MT) sets up two neural networks with identical architecture: a student model $F_S$ and a teacher model $F_T$. Given a batch $\mathbf{x}$ of training data, MT employs random augmentations $\eta_S, \eta_T$ to generate augmented inputs $\mathbf{x}_S$ and $\mathbf{x}_T$ for the student and teacher models correspondingly. During the feedforward pass, MT computes a weighted sum of a supervised classification loss and a consistency loss

$$\mathcal{L}_{MT} = \text{CE}(\mathbf{y}, \mathbf{f}_S^L) + \lambda_{\text{cons}} \text{MSE}(\mathbf{f}_S, \mathbf{f}_T), \quad (27)$$

where $\mathbf{f}_S, \mathbf{f}_T$ are posterior outputs from the student and teacher networks, $\mathbf{f}_S^L$ is the student's posterior output on the labeled data, and $\lambda_{\text{cons}}$ is a consistency weight hyperparameter. The classification loss is usually cross-entropy (CE), while the consistency loss is typically mean-squared error on the posteriors (MSE). The student model backpropagates directly using gradients from the loss $\mathcal{L}_{MT}$. In the meantime, the teacher model is updated via computing an exponential moving average (EMA) over the parameters of the student network. Recent papers have adapted MT for medical imaging tasks such as MR segmentation (Yu et al., 2019; Perone and Cohen-Adad, 2018) and nuclei classification (Su et al., 2019). We note that $\mathbf{f}_S, \mathbf{f}_T$ are similar to the views $\mathbf{f}_1, \mathbf{f}_2$ of NoT, except that MT uses the EMA update to compute $\mathbf{f}_T$, while NoT uses backpropagation to update both views.

Figure S2 illustrates the iterative process of co-training, a multi-view learning technique.

## S2. Additional Experiment Setup Details

We summarize the data split statistics for the NIH-14 Chest X-Ray, RSNA Brain CT and Knee MRNet datasets in Table S1. We also provide the hyperparameters used for the different labeling budgets, methods and datasets in our experiments in Table S2. Finally, we provide the adapted pipeline for the 3D MRNet classification task in Figure S3.

## S3. Additional Results: Comparisons to Baselines

For ease of assessment on quantitative results, we provide the detailed breakdown of AUROC scores for each dataset, labeling budget and method in Table S3. We also highlight comparisons to previously published state-of-the-art results where available. We note that, for the NIH-14 Chest X-Ray dataset, our fully-supervised baseline outperforms the numbers reported in the SRC-MT paper (Liu et al., 2020). We suspect that this is because we report the best results from either the trained model or its EMA copy. Also we highlight that the results of GraphX$^{\text{NET}}$ (Aviles-Rivero et al., 2019) and SRC-MT (Liu et al., 2020) are not directly comparable as they report metrics on a subset of the classes we considered and/or employ different backbones.

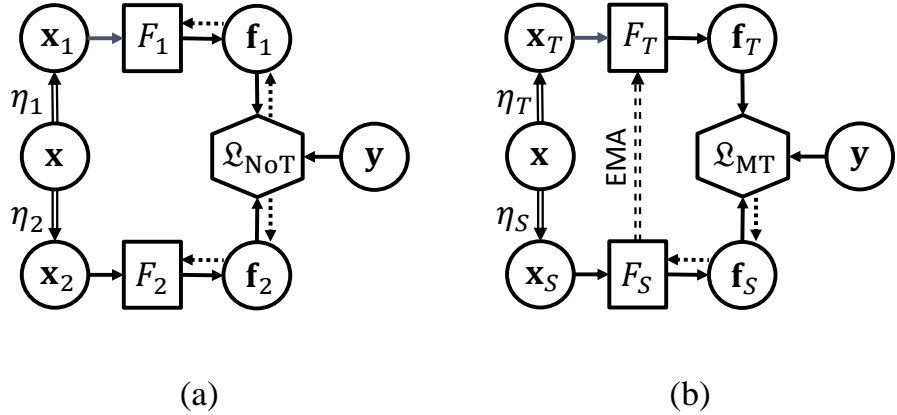

(a)                            (b)

Figure S1: **Training process of (a) MT and (b) NoT** on a batch of semi-supervised data. Solid and dotted arrows denote forward and backward passes, respectively. Double-line arrows denote random data augmentations, while double dashed arrow represents EMA update.

Table S1: Data statistics for the NIH-14, RSNA-CT and MRNet datasets.

| Dataset | $L_T : L_V$ | Split | Patients | Scans | Images |
|---|---|---|---|---|---|
| NIH-14 | 70 : 10 | Train | 21528 | 78468 | 78468 |
| | | Val. | 3090 | 11219 | 11219 |
| | | Test | 6187 | 22433 | 22433 |
| RSNA-CT | 60 : 20 | Train | 10247 | 10247 | 352839 |
| | | Val. | 3416 | 3416 | 117986 |
| | | Test | 3416 | 3416 | 117907 |
| MRNet | 64 : 16 | Train | 904 | 904 | 31156 |
| | | Val. | 226 | 226 | 7622 |
| | | Test | 120 | 120 | 4118 |

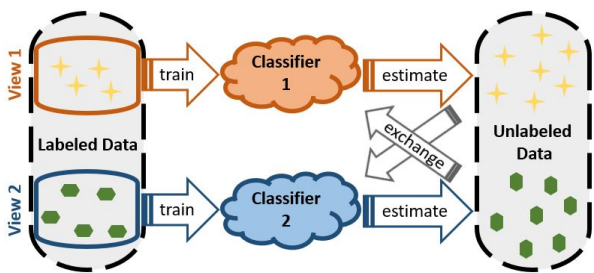

Figure S2: The iterative process of co-training in a two-view setup. This process continues until two views achieve a high level of agreement on unlabeled data.

### S4. Additional Results: NoT-GA Experiments

We have designed two other variants of the "DM-3311" setups, namely "DM-1133" and "DM-1313". First, the "DM-1133" has a class imbalance ratio of $\boldsymbol{\alpha}^L \propto [3, 3, 1, 1]$ on labeled data and a class imbalance ratio of $\boldsymbol{\alpha}^U \propto [1, 1, 3, 3]$ on unlabeled data, which is the reverse setup of "DM-3311". In contrast to the earlier setups, No Finding and Infiltration are now the classes with higher $\gamma$ values. Second, the "DM-1313" has a class imbalance ratio of $\boldsymbol{\alpha}^L \propto [3, 1, 3, 1]$ on labeled data and a class imbalance ratio of $\boldsymbol{\alpha}^U \propto [1, 3, 1, 3]$ on unlabeled data, which intentionally mixes up the (naturally) rare and common classes together. By comparing NoT against NoT-GA on these new setups, we further strengthen our understanding of the NoT-GA behaviors. The results are reported in Table S7, Table S8, Figure S4 and Figure S5.

Table S2: Hyperparameter tuning based on average validation AUROC. For VAT, multi-task KL divergence (KL$_{mt}$) offers > 10% AUROC boost. Other implementation details of NoT match with MT.

| | | NIH-14 | | | RSNA-CT | | | MRNet | | |
|---|---|---|---|---|---|---|---|---|---|---|
| | | Train | Val. | Test | Train | Val. | Test | Train | Val. | Test |
| **Labeling budget (%)** | | 1-5 | 10-20 | 25-100 | 0.25-0.5 | 1-2.5 | 5-100 | 1-10 | 15-25 | 30-100 |
| **MT** | $\alpha \in \{0.91, 0.93, \ldots, 0.99\}$ | 0.91 | 0.95 | 0.99 | 0.93 | | | 0.97 | 0.97 | 0.99 |
| | $\lambda_{cons} \in \{1, 2, \ldots, 196\}$ | 196 | | | 100 | | | 10 | | |
| **VAT** | $\epsilon \in \{1, 2, \ldots, 6\}$ | 2 | | | | | | 3 | | |
| | LDS $\in$ {KL, MSE, KL$_{mt}$} | KL$_{mt}$ | | | | | | | | |
| **NoT** | $\sigma_1^2 = \sigma_2^2 = 2^{-2}$ | | | | | | | | | |
| | $\sigma_y^2 \in \{2^{-2}, 2^{-3} \ldots, 2^{-7}\}$ | $2^{-2}$ | | | | | | $2^{-7}$ | $2^{-6}$ | |
| Early stopping | | 15 | 7 | 3 | 15 | 7 | 3 | 11 | 8 | 5 |
| Reduce learning rate patience | | 5 | 3 | 1 | 5 | 3 | 1 | 5 | 4 | 3 |
| Min no. of validation samples | | 113 | | | 268 | | | 25 | | |

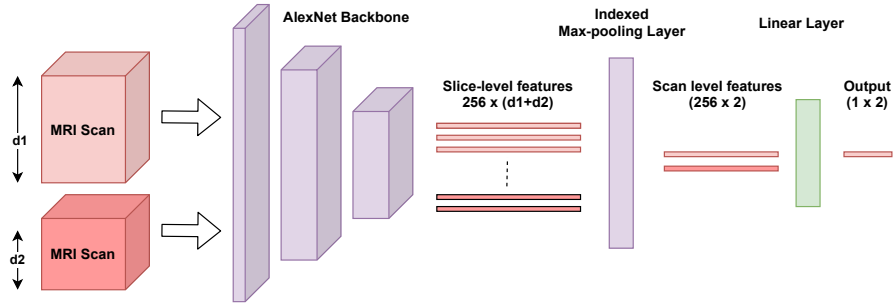

Figure S3: Adapted MR-Net Training Pipeline. As the MRI scans have variable slice depths, the original MR-Net can process only one scan at a time. We extend this architecture using indexed max-pooling, where the maximization is performed on the depth dimension of the scan. This allows us to process a batch of scans at a time.

### S5. Additional Results: MT vs. NoT

We provide additional plots from various seeds for the disagreement count comparison between MT and NoT. Figure S6 shows results on NIH-14 and Figure S7 shows results on RSNA Brain CT.

Table S3: Average AUROC scores (multiplied by 100) vs. labeling budget (%) for NIH-14 (top) and RSNA Brain CT (center) and Knee MRNet (bottom). At the best case budgets (bolded) of 5% in NIH-14, 1% in RSNA-CT and 25% in MRNet, NoT has 2.5%, 2.2% and 1.2% higher AUPRC than other SSL methods respectively. We include the average AUROC scores reported for SRC-MT (Liu et al., 2020) and GraphX$^{NET}$ (Aviles-Rivero et al., 2019).

| NIH-14 | | | | | | | |
|---|---|---|---|---|---|---|---|
| **Budget (Images)** | **SUP** | **PSU** | **VAT** | **MT** | **NoT** | **GraphX$^{NET}$** | **SRC-MT** |
| 1.3 (1177) | 66.53 ± 0.38 | 66.48 ± 0.43 | 69.56 ± 0.17 | 68.97 ± 0.63 | **70.69 ± 0.15** | — | — |
| 2 (1569) | 66.85 ± 1.45 | 67.64 ± 0.70 | 69.63 ± 0.23 | 70.42 ± 0.58 | **72.60 ± 0.18** | 53.00 | 66.95 |
| **5 (3923)** | 70.68 ± 1.31 | 70.93 ± 0.99 | 73.94 ± 0.14 | 73.60 ± 0.70 | **77.04 ± 0.22** | 58.00 | 72.29 |
| 10 (7846) | 75.69 ± 1.17 | 76.00 ± 0.67 | 77.15 ± 1.06 | 76.98 ± 0.02 | **77.61 ± 0.54** | 63.00 | 75.28 |
| 20 (15693) | 77.19 ± 0.75 | 78.06 ± 0.47 | 79.38 ± 0.10 | 78.66 ± 0.64 | **79.49 ± 0.89** | 78.00 | 79.23 |
| 50 (39234) | 80.57 ± 0.68 | 81.46 ± 0.34 | 82.01 ± 0.14 | 81.78 ± 0.11 | **82.10 ± 0.05** | — | — |
| **100 (78468)** | **83.33 ± 0.38** | — | — | — | — | — | 81.75 |
| RSNA-CT | | | | | | | |
| **Budget (Slices)** | **SUP** | **PSU** | **VAT** | **MT** | **NoT** | **GraphX$^{NET}$** | **SRC-MT** |
| 0.25 (749) | 70.77 ± 3.52 | 72.19 ± 1.81 | 71.07 ± 2.72 | 74.86 ± 2.23 | **76.53 ± 1.84** | — | — |
| 0.5 (1777) | 80.55 ± 0.97 | 80.89 ± 1.43 | 82.50 ± 0.60 | 82.15 ± 1.43 | **83.57 ± 0.98** | — | — |
| **1 (3495)** | 80.01 ± 1.19 | 81.53 ± 0.32 | 81.41 ± 2.17 | 80.90 ± 1.91 | **84.50 ± 1.86** | — | — |
| 2.5 (6744) | 86.13 ± 1.10 | 87.01 ± 0.15 | 87.22 ± 0.40 | 86.26 ± 0.58 | **89.81 ± 0.23** | — | — |
| 5 (17242) | 91.31 ± 0.29 | **91.84 ± 0.37** | 90.53 ± 0.38 | 91.24 ± 0.18 | 91.79 ± 0.73 | — | — |
| 10 (33560) | 92.74 ± 0.38 | 93.27 ± 0.03 | 92.84 ± 0.44 | 92.62 ± 0.64 | **93.31 ± 0.80** | — | — |
| **100 (352839)** | **96.69 ± 0.11** | — | — | — | — | — | — |
| MRNet | | | | | | | |
| **Budget (Scans)** | **SUP** | **PSU** | **VAT** | **MT** | **NoT** | **GraphX$^{NET}$** | **SRC-MT** |
| 5 (32) | 66.26 ± 3.13 | 67.31 ± 3.72 | 71.37 ± 2.42 | 68.36 ± 2.70 | **73.25 ± 2.78** | — | — |
| 7.5 (55) | 68.51 ± 6.09 | 73.81 ± 3.57 | 75.38 ± 5.23 | 73.33 ± 2.27 | **77.54 ± 2.11** | — | — |
| 15 (137) | 83.72 ± 2.08 | 85.63 ± 1.27 | 85.75 ± 2.85 | 87.12 ± 0.44 | **88.35 ± 1.38** | — | — |
| **25 (227)** | 85.07 ± 5.50 | 85.94 ± 5.39 | 85.12 ± 3.12 | 85.15 ± 3.86 | **89.19 ± 0.89** | — | — |
| 50 (452) | 89.81 ± 1.08 | 90.47 ± 1.31 | 90.23 ± 0.47 | 90.89 ± 1.72 | **91.49 ± 0.90** | — | — |
| **100 (904)** | **91.03 ± 0.33** | — | — | — | — | — | — |

* For GraphX$^{NET}$ (Aviles-Rivero et al., 2019), the average AUROC scores were reported from 8 labels, namely Atelectasis, Cardiomegaly, Effusion, Infiltration, Mass, Nodule, Pneumonia and Pneumothorax, whereas the rest of the methods were averaged over all of the 14 labels.

* While the rest of methods use a DenseNet-121 backbone, the SRC-MT reports its results using a DenseNet-169 backbone and the GraphX$^{NET}$ utilizes graph based representation.

Table S4: Validation AUC while training on 5% annotation budget on the Chest-XRay14 dataset for the selection of best loss function

|  | LDS divergence function | | |
|---|---|---|---|
| eps | MSE | KL Multiclass | KL |
| 2 | 0.7425 | **0.7503** | 0.6532 |
| 4 | 0.7436 | 0.7499 | 0.5297 |
| 6 | 0.7269 | 0.7450 | 0.6510 |
| 8 | 0.7484 | 0.7472 | 0.5346 |
| 10 | 0.7393 | 0.7457 | 0.5459 |

Table S5: **Class Distribution Mismatch**: Average Per-Class AUPRC scores (multiplied by 100)

| **Methods** | **No Finding** | **Infiltration** | **Pneumothorax** | **Mass** | **Average** |
|---|---|---|---|---|---|
| SUP | 61.18 ± 0.66 | 44.58 ± 1.48 | 33.44 ± 2.37 | 19.99 ± 2.77 | 39.80 ± 1.82 |
| MT | 60.93 ± 1.53 | 45.31 ± 1.18 | 35.58 ± 1.41 | 22.91 ± 3.92 | 41.18 ± 2.01 |
| VAT | 61.58 ± 0.88 | 45.10 ± 1.02 | 34.75 ± 0.87 | 20.74 ± 2.54 | 40.54 ± 1.33 |
| NoT | 62.30 ± 0.83 | 48.27 ± 0.61 | 41.71 ± 1.25 | 34.81 ± 3.17 | 46.77 ± 1.47 |
| NoT-GA | **62.79 ± 0.71** | **48.83 ± 0.52** | **42.33 ± 1.13** | **38.66 ± 2.75** | **48.15 ± 1.28** |
| Fully Supervised | 65.51 ± 1.22 | 55.05 ± 0.88 | 46.30 ± 0.70 | 41.44 ± 1.05 | 52.08 ± 0.96 |

Table S6: Results on DM-3311 Setup

| **Metric** | **Method** | **No Finding** | **Infiltration** | **Pneumothorax** | **Mass** | **Average** |
|---|---|---|---|---|---|---|
| AUPRC | NoT | 51.25 ± 0.68 | 53.39 ± 0.68 | 50.04 ± 1.26 | 46.7 ± 1.89 | 50.34 ± 1.13 |
| | NoT-GA | 51.07 ± 0.45 | 53.36 ± 0.77 | 53.28 ± 0.74 | 60.58 ± 2.75 | 54.57 ± 1.18 |
| Precision (threshold 0.5) | NoT | 0.585 ± 0.033 | 0.638 ± 0.081 | 0.283 ± 0.021 | 0.216 ± 0.028 | – |
| | NoT-GA | 0.517 ± 0.01 | 0.548 ± 0.002 | 0.431 ± 0.028 | 0.527 ± 0.053 | – |
| Recall (threshold 0.5) | NoT | 0.16 ± 0.076 | 0.115 ± 0.082 | 0.803 ± 0.021 | 0.792 ± 0.034 | – |
| | NoT-GA | 0.497 ± 0.03 | 0.419 ± 0.05 | 0.705 ± 0.018 | 0.619 ± 0.025 | – |

Table S7: Results on DM-1133 Setup

| Metric | Method | No Finding | Infiltration | Pneumothorax | Mass | Average |
|---|---|---|---|---|---|---|
| AUPRC | NoT | 28.31 ± 0.83 | 30.72 ± 1.63 | 70.75 ± 1.15 | 64.48 ± 2.18 | 48.57 ± 1.45 |
| | NoT-GA | 26.67 ± 0.45 | 30.68 ± 1.12 | 72.66 ± 0.81 | 67.02 ± 1.45 | 49.26 ± 0.96 |
| Precision (threshold 0.5) | NoT | 0.174 ± 0.018 | 0.177 ± 0.027 | 0.788 ± 0.034 | 0.886 ± 0.069 | - |
| | NoT-GA | 0.239 ± 0.024 | 0.315 ± 0.025 | 0.693 ± 0.022 | 0.714 ± 0.051 | - |
| Recall (threshold 0.5) | NoT | 0.606 ± 0.054 | 0.541 ± 0.054 | 0.279 ± 0.123 | 0.090 ± 0.058 | - |
| | NoT-GA | 0.497 ± 0.048 | 0.4358 ± 0.047 | 0.614 ± 0.019 | 0.430 ± 0.087 | - |

Table S8: Results on DM-1313 Setup

| Metric | Method | No Finding | Infiltration | Pneumothorax | Mass | Average |
|---|---|---|---|---|---|---|
| AUPRC | NoT | 23.03 ± 1.03 | 55.7 ± 1.68 | 47.51 ± 3.02 | 57.99 ± 6.84 | 46.06 ± 3.15 |
| | NoT-GA | 25.97 ± 0.62 | 58.57 ± 1.42 | 55.59 ± 0.68 | 63.76 ± 1.93 | 50.97 ± 1.16 |
| Precision (threshold 0.5) | NoT | 0.161 ± 0.008 | 0.552 ± 0.311 | 0.259 ± 0.015 | 0.641 ± 0.360 | - |
| | NoT-GA | 0.203 ± 0.022 | 0.618 ± 0.033 | 0.396 ± 0.028 | 0.640 ± 0.018 | - |
| Recall (threshold 0.5) | NoT | 0.700 ± 0.024 | 0.035 ± 0.021 | 0.823 ± 0.052 | 0.082 ± 0.053 | - |
| | NoT-GA | 0.549 ± 0.062 | 0.303 ± 0.047 | 0.781 ± 0.031 | 0.385 ± 0.051 | - |

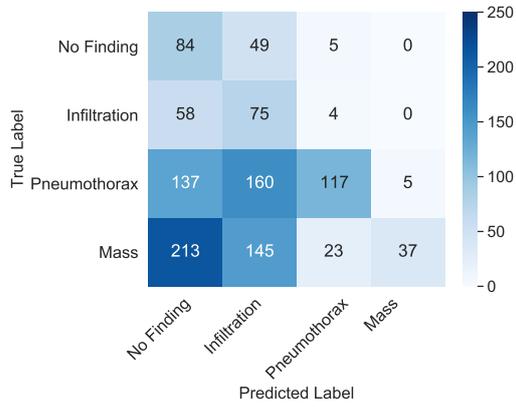 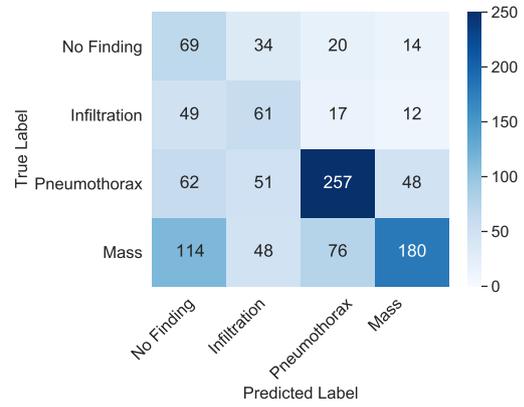

(a) NoTeacher (NoT)  (b) NoTeacher with Generative Assumption (NoT-GA)

Figure S4: **Results on DM-1133 Setup**: Average confusion matrix (threshold 0.5) over 5 seeds with vanilla NoT (a) and with NoT-GA (b).

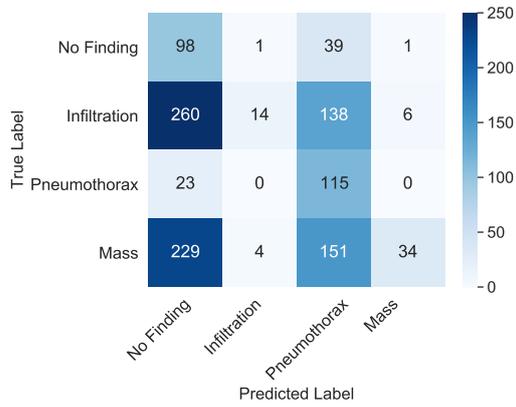 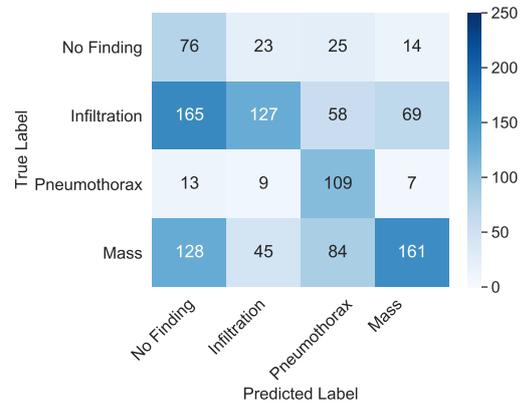

(a) NoTeacher (NoT)  (b) NoTeacher with Generative Assumption (NoT-GA)

Figure S5: **Results on DM-1313 Setup**: Average confusion matrix (threshold 0.5) over 5 seeds with vanilla NoT (a) and with NoT-GA (b).

Figure S6: Performance as a function of training as indicated by AUROC scores on the validation set (left vertical axis) and disagreement statistics (right vertical axis). Results are from NIH-14 Chest X-Ray with 5% labeling budget, $X = 40$, $\tau = 0.25$ with various seeds.

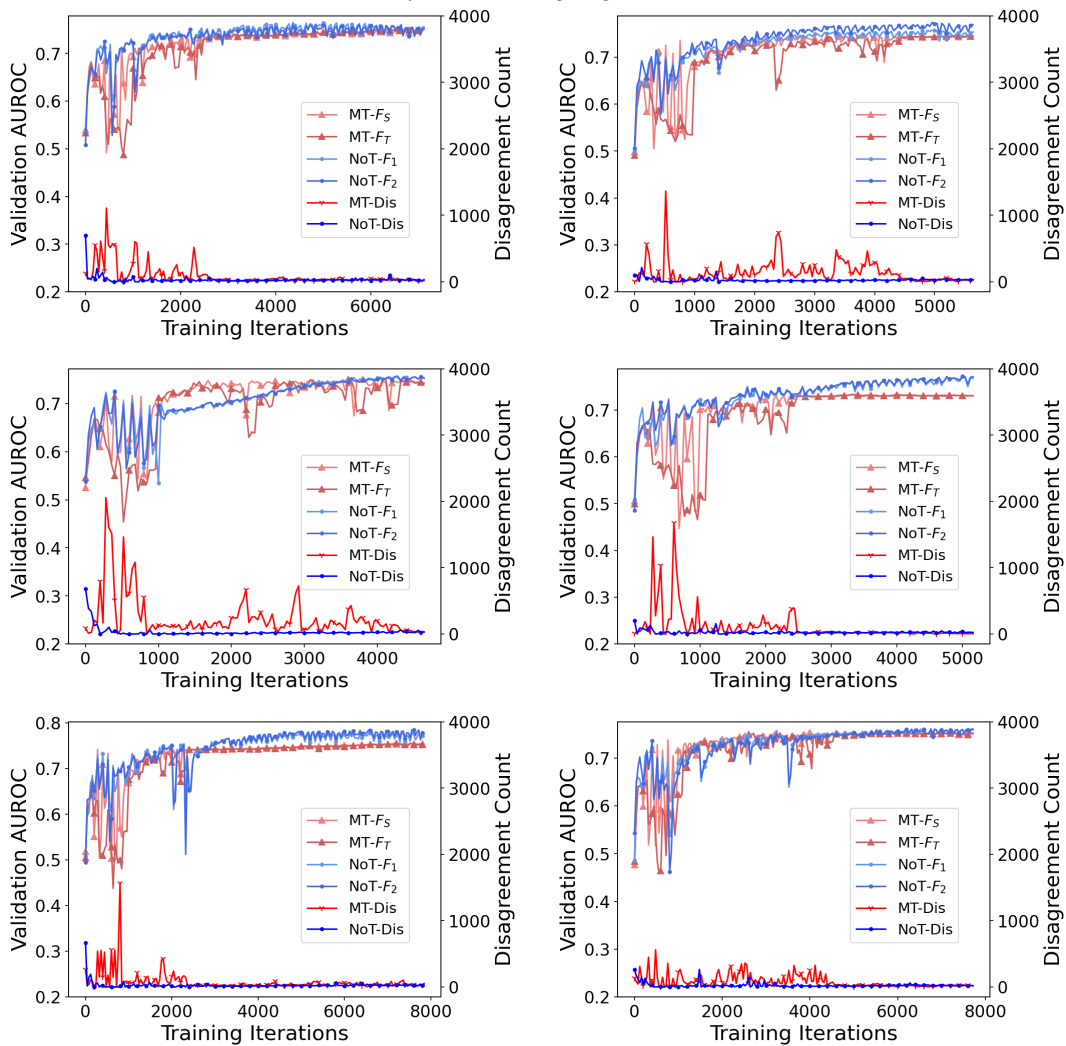

Figure S7: Performance as a function of training as indicated by AUROC scores on the validation set (left vertical axis) and disagreement statistics (right vertical axis). Results are from RSNA Brain CT with 2.5% labeling budget, $X = 40$, $\tau = 0.2$ with various seeds.

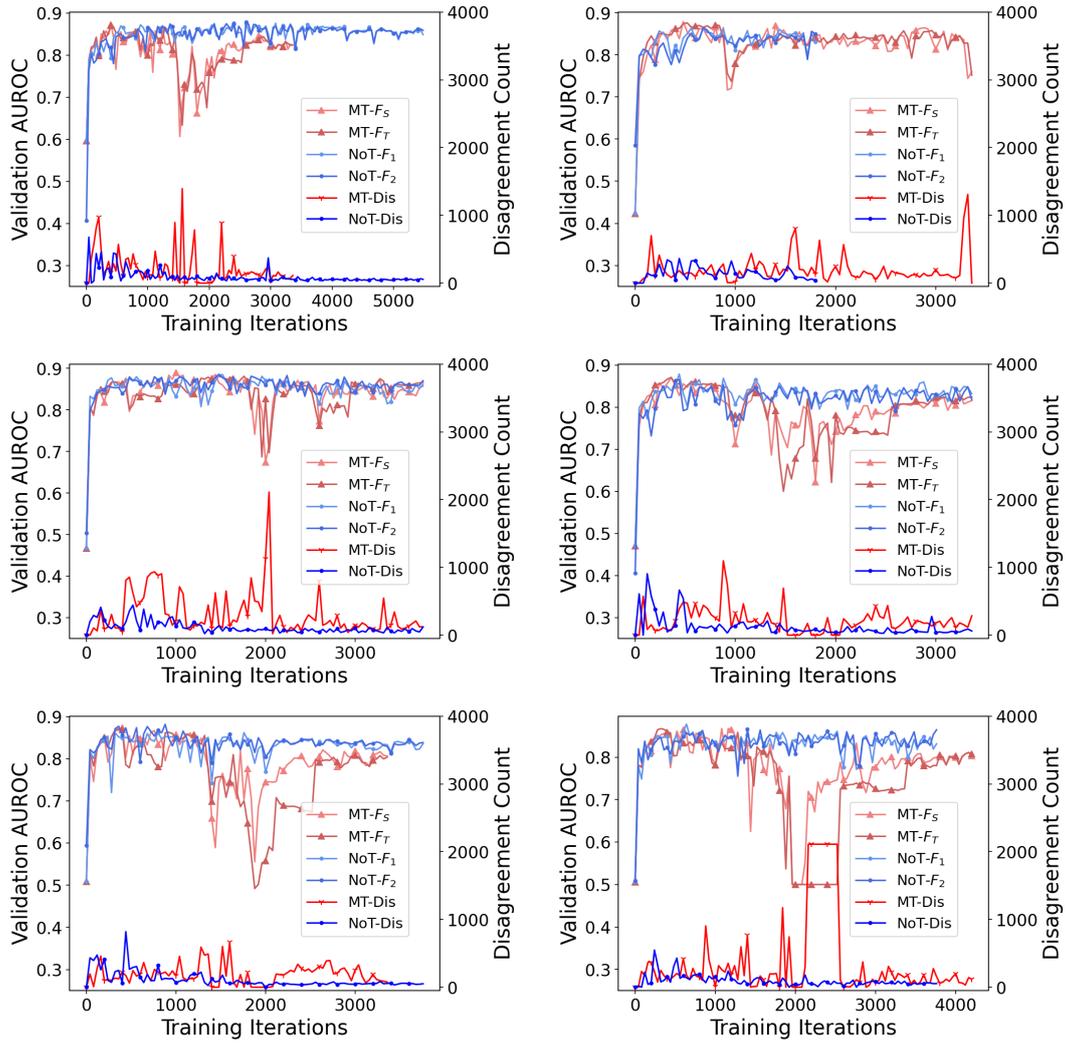